\documentclass[a4paper,twocolumn,preprint,5p,authoryear,sort&compress,8pt]{elsarticle}

\usepackage{stfloats}
\usepackage[T1]{fontenc}
\usepackage{lmodern}
\usepackage{lipsum}
\journal{NeuroImage}
\usepackage{lineno,hyperref}
\usepackage{amssymb}
\usepackage{multicol}

\usepackage{graphicx} 
\usepackage{bbold}
\usepackage{makeidx}  
\usepackage[misc]{ifsym}
\usepackage{amsmath}
\usepackage{multirow}
\usepackage{array}
\usepackage{siunitx}
\usepackage[dvipsnames]{xcolor}

\DeclareMathOperator*{\argmin}{arg\,min}
\DeclareMathOperator{\Tr}{Tr}
\newcommand{\rpm}{\raisebox{.2ex}{$\scriptstyle\pm$}}
\modulolinenumbers[5]
\usepackage{mathtools}

\biboptions{semicolon}

\begin{document}

\begin{frontmatter}

\title{Deep sr-DDL: Deep Structurally Regularized Dynamic Dictionary \\ Learning to Integrate Multimodal and Dynamic Functional \\ Connectomics  data  for Multidimensional Clinical Characterizations}
\author[wse]{N.S.~D'Souza\corref{cor1}}
\ead{Shimona.Niharika.Dsouza@jhu.edu}
\author[KKI,Neu]{M.B.~Nebel}
\author[KKI]{D.~Crocetti}
\author[KKI]{J.~Robinson}
\author[KKI,Neu]{N.~Wymbs}
\author[KKI,Neu,Psych]{S.H.~Mostofsky}
\author[wse]{A.~Venkataraman}

\cortext[cor1]{Corresponding author}

\address[wse]{Department of Electrical and Computer Engineering, Johns Hopkins University, USA}
\address[KKI]{Center for Neurodevelopmental \& Imaging Research, Kennedy Krieger Institute, USA}
\address[Neu]{Department of Neurology, Johns Hopkins School of Medicine, USA}
\address[Psych]{Department of Psychiatry and Behavioral Science, Johns Hopkins School of Medicine, USA}

\begin{abstract}
We propose a novel integrated framework that jointly models complementary information from resting-state functional MRI (rs-fMRI) connectivity and diffusion tensor imaging (DTI) tractography to extract biomarkers of brain connectivity predictive of behavior. Our framework couples a generative model of the connectomics data with a deep network that predicts behavioral scores. The generative component is a structurally-regularized Dynamic Dictionary Learning (sr-DDL) model that decomposes the dynamic rs-fMRI correlation matrices into a collection of shared basis networks and time varying subject-specific loadings. We use the DTI tractography to regularize this matrix factorization and learn anatomically informed functional connectivity profiles. The deep component of our framework is an LSTM-ANN block, which uses the temporal evolution of the subject-specific sr-DDL loadings to predict multidimensional clinical characterizations. Our joint optimization strategy collectively estimates the basis networks, the subject-specific time-varying loadings, and the neural network weights. We validate our framework on a dataset of neurotypical individuals from the Human Connectome Project (HCP) database to map to cognition and on a separate multi-score prediction task on individuals diagnosed with Autism Spectrum Disorder (ASD) in a five-fold cross validation setting. Our hybrid model outperforms several state-of-the-art approaches at clinical outcome prediction and learns interpretable multimodal neural signatures of brain organization.
\end{abstract}


\end{frontmatter}

\section{Introduction}
\begin{figure*}[t!]
   \centering
   \includegraphics[scale=0.55]{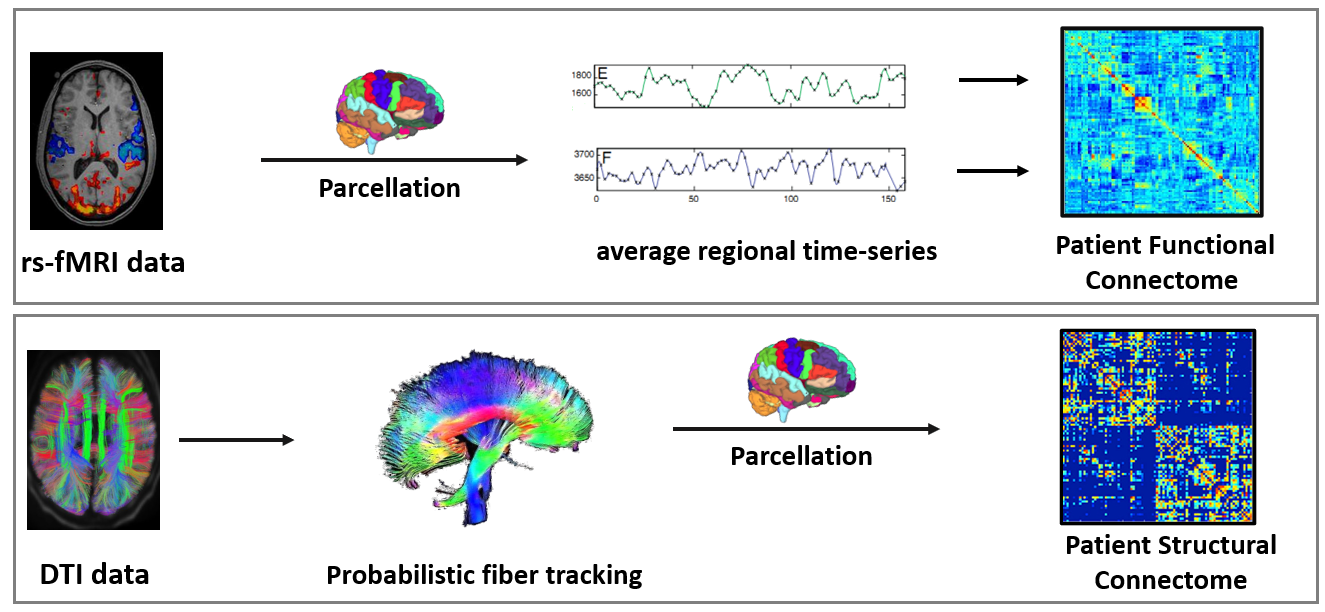}
  \caption{ \textbf{Top:} For the fMRI data, we group voxels in the brain into ROIs defined by a standard atlas and compute the average time courses for each ROI. The correlation matrix captures the synchrony in the average time courses. 
  \textbf{Bottom} Tractography is performed on the raw DWI data to track the path of neuronal fibers in the brain. Based on the parcellation scheme, we construct a map of the fibre tracts between ROIs in the brain. The same parcellation scheme is used for both modalities. 
  }\label{Ex}
\end{figure*}

Functional magnetic resonance imaging (fMRI) quantifies the changes in blood flow and oxygenation in the regions associated with neuronal activity. More specifically, resting state fMRI (rs-fMRI) is acquired in the absence of a task paradigm, thus allowing us to probe the spontaneous co-activation patterns in the brain. It is believed that the co-activations reflect the intrinsic functional connectivity between brain regions [\cite{fox2007spontaneous}]. In contrast to fMRI, Diffusion Tensor Imaging (DTI) [\cite{assaf2008diffusion}] assesses structural connectivity by measuring the diffusion of water molecules across neuronal fibres in the brain. Going one step further, we can use tractography to construct detailed $3D$ maps of anatomical pathways within the brain based on the diffusion tensors. There is strong evidence in literature of the correspondence between functional and structural pathways within the brain [\cite{skudlarski2008measuring}], with several studies suggesting that this functional connectivity may be mediated by either direct or indirect anatomical connections [\cite{fukushima2018structure,bowman2012determining,honey2009predicting,atasoy2016human}]. Thus, rs-fMRI and DTI data provide complementary information about function and structure respectively, which when integrated together can be used to construct a more comprehensive  view of brain organization both in health and disease. As a result, multimodal integration has become an important topic of study for the characterization of neuropsychiatric disorders such as Autism Spectrum Disorder (ASD) [\cite{vissers2012brain}], Attention Deficit Hyperactivity Disorder (ADHD) [\cite{weyandt2013neuroimaging}], and Schizophrenia [\cite{niznikiewicz2003recent}].

\par Traditional multimodal analyses of rs-fMRI and DTI data have largely focused on post-hoc statistical comparisons of features extracted from the data. For example, simple statistical differences in rs-fMRI and DTI connectivity between subjects have been used to discover disrupted patterns of brain organization in Alzheimer's disease [\cite{hahn2013selectively}] and Progressive Supranuclear Palsy (PSP) [\cite{whitwell2011disrupted}]. On a population level, classical multivariate analysis [\cite{goble2012neural} \cite{andrews2007disruption}] or random effects models [\cite{propper2010combined}] are employed to independently compute and then combine features from both modalities. Despite their past success at biomarker discovery, these techniques often fail to generalize at a patient-specific level. Furthermore, they often ignore higher-order interactions between multiple subsystems in the brain, which is known to be critical for understanding complex neuropsychiatric disorders  [\cite{kaiser2010neural,koshino2005functional}]. These shortcomings have paved the way for the development of the network based view of brain connectivity that simultaneously accounts for both inter-subject and intra-subject variability.

\par In the case of fMRI, network-based models often group voxels in the brain into regions of interest (ROIs) using a standard anatomical or functional atlas. Next, the functional relationships between these regions are determined based on the synchrony between representative (often average) regional time series. This information is typically represented in terms of a static functional connectivity matrix as shown in Fig.~\ref{Ex} (top). In case of DTI, tractography is used to estimate the fiber tracts between the ROIs in the brain from the voxel-level diffusion tensors, from which features such as the anisotropy or the number of fibers can be extracted. Similar to the functional connectome, the structural connectivity matrix captures the strength of the pairwise anatomical connection between different ROIs, as seen in Fig.~\ref{Ex} (bottom). 

\par Some of the simplest approaches to analyzing network properties borrow heavily from the field of graph theory. For example, the works of [\cite{sporns2004organization,rubinov2010complex,bullmore2009complex}] use aggregate network measures, such as node degree, betweenness centrality, and eigenvector centrality to study the organization of the brain. These measures compactly summarize the connectivity information onto a restricted set of nodes that can be mapped back to the brain. A more global network property is small-worldedness [\cite{bassett2006small}], which describes an architecture of sparsely connected clusters of nodes. Complementary changes in small-worldedness in both anatomical and functional networks have been well documented across the literature [\cite{park2008comparison,sun2014disrupted}], with concurrent disruptions of functional networks [\cite{wang2009functional}] or structural networks [\cite{wang2012anatomical}] implicated in neuropsychiatric disorders such as schizophrenia. The main limitation of these approaches is that they independently analyze the  fMRI and DTI data, and as such, draw heuristic conclusions about the relationship between the two modalities.

\par Community detection techniques have been widely used for understanding the organization of complex systems such as the brain [\cite{bardella2016hierarchical,nandakumar2018defining}]. Other examples include the work of [\cite{venkataraman2013connectivity}] that identifies abnormal connectivity in schizophrenia, and  [\cite{venkataraman2016bayesian}], which characterizes the social and communicative deficits associated with autism. An alternative network topology is the hub-spoke model, used by [\cite{venkataraman2013connectivity}, \cite{venkataraman2012brain}, \cite{venkataraman2015unbiased}], that targets regions associated with a large number of altered rs-fMRI connections. These methods, however, exclusively focus on functional connectivity and do not incorporate structure. In this light, the work of [\cite{venkataraman2011joint}] proposes a probabilistic framework that jointly models latent anatomical and functional connectivity to discover population-level differences in schizophrenia. Similarly, the work of [\cite{higgins2018integrative}] uses a unified Bayesian framework to identify gender-differences in multimodal connectivity patterns across different age groups. While successful at combining multi-modal information for group differentiation, these techniques do not directly address inter-individual variability.

\par Data-driven methods integrating structural and functional connectivity focus heavily on groupwise discrimination from the static connectomes. These methods usually follow a two-step approach where feature selectors and discriminators are trained sequentially in a pipeline. For example, the authors in [\cite{wee2012identification}] combine graph theoretic features computed from rs-fMRI and DTI graphs with Support Vector Machines (SVMs) to identify individuals with Mild Cognitive Impairment. Another example is the work of [\cite{sui2013combination}], which employs a pipeline consisting of joint-Independent Component Analysis (j-ICA) on the two modalities followed by Canonical Correlation Analysis (CCA) to combine them and distinguish schizophrenia patients from controls. In contrast to the pipelined approaches, end-to-end deep learning methods combining feature selection and prediction are becoming ubiquitous in neuroimaging studies. These are highly successful due to their ability to learn complex abstractions directly from input data. As an example, the work of [\cite{aghdam2018combination}] uses a Deep Belief Network (DBN) on multimodal data to disambiguate patients with Autism Spectrum Disorder from healthy controls. However, none of the above methods tackle continuous-valued prediction, for example, quantifying a continuous level of deficit. 

\par{In the continuous prediction realm, our previous works in [\cite{d2018generative,d2020joint}] and [\cite{d2019coupled}] combine dictionary learning on rs-fMRI correlation matrices with linear, non-linear regression models respectively to predict a single measure of clinical severity. These methods combine the rs-fMRI representation with the prediction in a coupled optimization framework. While they use a similar coupled optimization strategy, they fail to generalize to predicting multiple deficits (i.e. multi-score prediction). On the other hand, recent works of [\cite{kawahara2017brainnetcnn,d2019integrating}] have demonstrated the power of deep neural networks to map to multiple clinical/cognitive outcomes from rs-fMRI and DTI data separately. While promising, all of these methods focus on a single neuroimaging modality and do not exploit complementary interactions between structural and functional connectivity. In addition, the aforementioned techniques rely on static rs-fMRI correlation matrices as input. Consequently, they largely ignore the dynamics of evolution of the functional scan.}

\begin{figure*}[t!]
   \centering
   \includegraphics[scale=0.55]{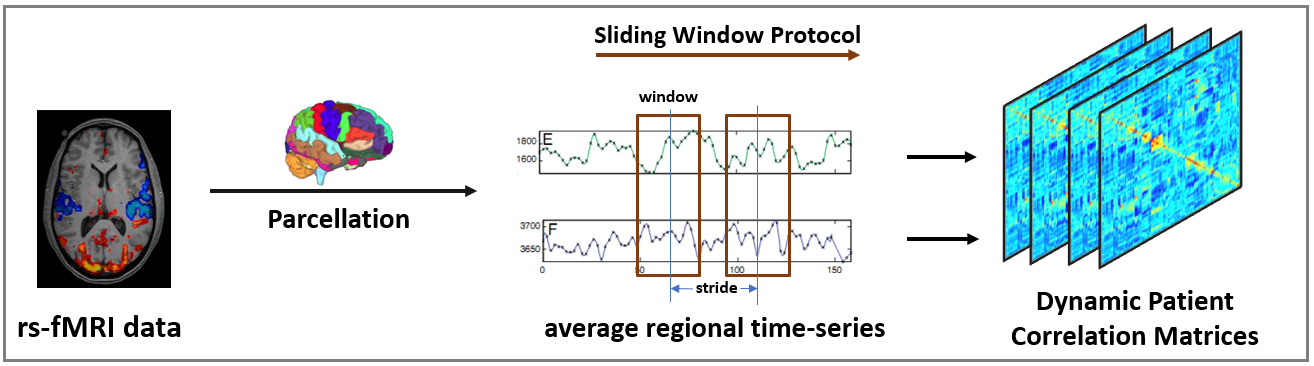}
   \caption{First, the ROI's defined by a standard atlas are used to compute regional time series. Then, a sliding window protocol defined by window length and stride is applied to extract the dynamic patient correlation matrices. As in the static case, the dynamic matrices measure the synchrony between regional time series, but as a function of time.}\label{Ex_Dyn}
\end{figure*}

\par There is now growing evidence that functional connectivity is a dynamic process that toggles between different intrinsic states evolving over a static structural connectome [\cite{cabral2017functional}].  These states manifest over short time windows that are typically of the order of a tens of seconds to a few minutes. Several studies such as [\cite{rashid2014dynamic,price2014multiple,nandakumar2020multi}] indicate the importance of modeling this evolution for characterizing neuropsychiatric disorders such as schizophrenia and Autism Spectrum Disorder (ASD). The dynamic connectivity among ROIs in the brain is typically captured via a sliding window protocol, defined by the window length and stride, as illustrated in Fig.~\ref{Ex_Dyn}. The window length defines the length of the time sequence considered by each dynamic correlation matrix, while the stride controls the overlap in successive sliding windows. Recently, model based alternatives that detect dynamic changes in correlation between large-scale brain networks such as the Default Mode Network, Somatosensory Network etc have been developed. An example is the Dynamic Conditional Correlation (DCC) protocol that was initially developed in the econometrics and finance literature [\cite{engle2002dynamic}] and later adapted to the study of brain organization using rs-fMRI [\cite{lindquist2016dynamic}]. It poses a time-varying matrix estimation problem to explicitly model the evolution of connectivity patterns in the brain, and has shown robustness in the test-retest setting [\cite{lindquist2014evaluating}] with rs-fMRI. Unfortunately, this method is unstable when scaled up [\cite{caporin2013ten,aielli2013dynamic}], for example to a whole brain ROI-level analysis of dynamic connectivity, likely due to ill conditioning of the correlation matrices in the absence of additional regularization. Consequently, most dynamic connectivity studies continue to rely on sliding-window correlations as inputs. Examples include [\cite{cai2017estimation}], where the authors use a sparse decomposition of the rs-fMRI connectomes, or [\cite{rabany2019dynamic}], which employs a temporal clustering for ASD/control discrimination. Nevertheless, these approaches focus exclusively on rs-fMRI and completely ignore structural information. 

\par {We propose a deep-generative hybrid model, i.e. the deep sr-DDL, that integrates structural and dynamic functional connectivity with behavior into a unified optimization framework.}

\subsection{{Our Contribution}}
{The contributions of this work are two-fold. From an application standpoint, we develop a unified framework to integrate structural (DTI) and dynamic rs-fMRI connectivity together with behavior. From a technical standpoint, we propose a unique alternative to black-box deep learning methods by combining the interpretability of classical techniques with the representational power of strategically-designed deep neural networks. As a starting point, we leverage the dictionary learning frameworks of [\cite{eavani2015identifying}; \cite{d2018generative}; \cite{d2019coupled,d2019integrating}], which extract group-level subnetworks from static rs-fMRI correlation matrices. Our deep sr-DDL carries this method further via two main components:}

{
\begin{itemize}
    \item A generative dictionary learning component to represent the multimodal and dynamic data
    \item A deep network to model the temporal trends and predict behavioral scores.
\end{itemize} }

\par {Our generative component is a structurally regularized Dynamic Dictionary Learning (sr-DDL), which uses a DTI tractography prior to regularize a matrix factorization of the dynamic rs-fMRI correlation matrices. The sr-DDL decomposes dynamic rs-fMRI correlation matrices into a collection of shared bases, and time-varying subject specific loadings. These loadings are input to a deep network which is comprised of a Long-Short Term Memory (LSTM) module to model temporal trends and an ANN that predicts clinical scores. The key to this generative-deep hybrid is our} {coupled optimization procedure} {, which jointly estimates the bases, loadings, and neural network weights most predictive of the individual behavioral profile.}
\par {A preliminary version of our work was published in MICCAI 2020 [\cite{d2020deep}]. In this journal, we provide a detailed analysis of our framework where we validate on both synthetic data and two separate real-world datasets. The first of these includes a subset of healthy adults from the publicly available Human Connectomme Project (HCP) [\cite{van2012human}]. This helps us evaluate the efficacy of our framework at predicting cognitive outcomes from the rs-fMRI and DTI scans. Next, we examine a a clinical dataset consisting of children diagnosed with Autism Spectrum Disorder (ASD). The presentation of ASD is known to be heterogeneous with individuals exhibiting a wide spectrum of behavioral impairments in terms of social reciprocity, communicative functioning, and repetitive/restrictive behaviours [\cite{spitzer1980diagnostic}], quantified via clinical severity measures. We observed that our method outperforms several state-of-the-art approaches at predicting behavioral performance in unseen individuals from their connectomics data for both datasets. This illustrates that our method is reproducible. Furthermore, we provide a detailed presentation of our clinical results, especially the subnetworks identified by the model in both datasets. We conclude with a discussion on the generalizability, and robustness and potential directions of future work. }

\par In summary, our joint objective balances generalizability with interpretability, bridging the representational gap between structure, function and behavior. Our experiments highlight the potential of our deep sr-DDL framework for providing a more holistic view of neuropsychiatric diseases.

\section{Materials and Methods }
\subsection{A Deep Generative Hybrid Model to integrate Multimodal and Dynamic Connectivity with Behavior }
\begin{figure*}[t!]
   \centering
   \includegraphics[scale=0.55]{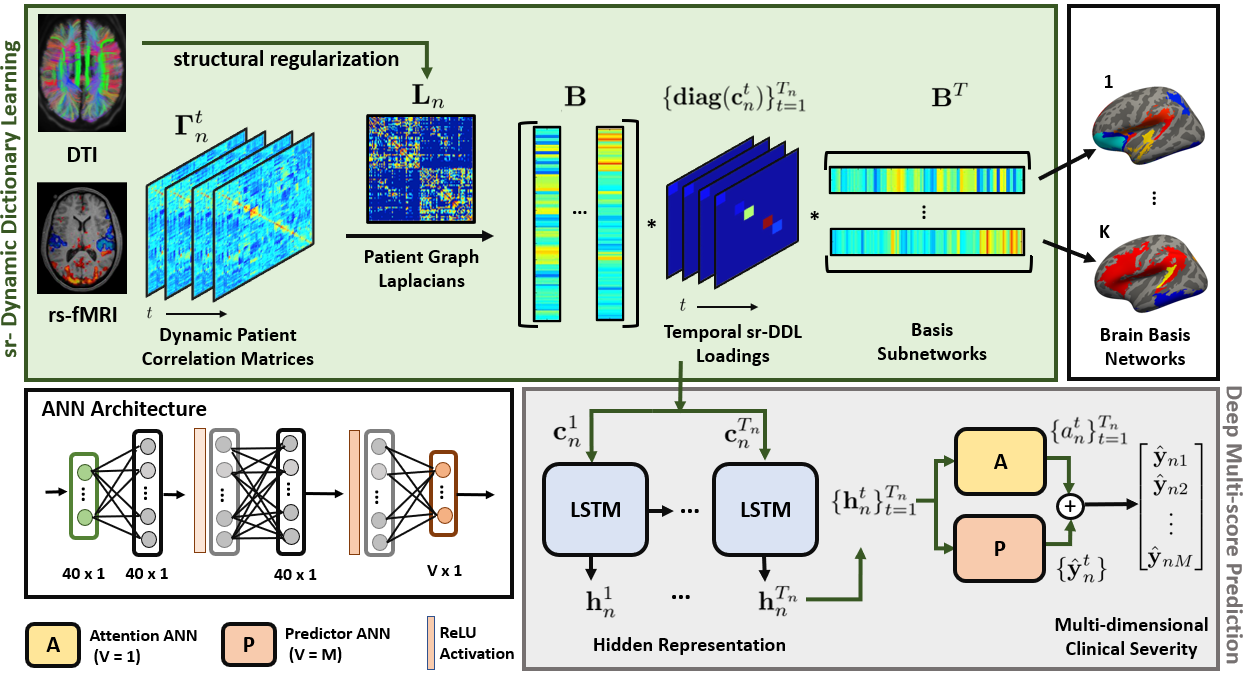}
   \caption{Framework to integrate structural and dynamic functional connectivity for clinical severity prediction \textbf{Green Box:}~The generative sr-DDL module. The rs-fMRI dynamic correlation matrices are decomposed into the subnetwork basis and time-varying subject-specific loadings. The DTI connectivity regularizes this decomposition. \textbf{Purple Box:}~Deep LSTM-ANN module for multi-score prediction. The sr-DDL coefficients are input into the LSTM to generate a hidden representation. The predictor ANN (P-ANN) generates a time varying estimate for the scores, while the attention ANN (A-ANN) weights the predictions across time to generate the final clinical severity estimate.}  \label{CCS}
\end{figure*}
Fig.~\ref{CCS} presents a graphical overview of our framework. {We have two sets of inputs to the model for each individual namely, the dynamic individual-specific correlation matrices, and the DTI structural connectome graph (upper left). Our outputs are the scalar clinical scores (bottom right).} We use the sliding window approach in Fig.~\ref{Ex_Dyn} to extract dynamic rs-fMRI correlation matrices and tractography to extract the DTI connectomes as shown in Fig.~\ref{Ex}. The DTI input to our model is the Graph Laplacian obtained from a binary DTI adjacency matrix capturing the presence/absence of a fiber between regions. Finally, the behavioral scores for each individual are obtained from an expert assessment. This score can correspond to either cognitive outcomes or severity of symptoms in case of neurodevelopmental diseases.
\par The green box in Fig.~\ref{CCS} describes the generative component of our framework. Here, the dynamic rs-fMRI correlation matrices are decomposed using a structurally regularized dynamic dictionary learning (sr-DDL). The columns in the bases subnetworks capture representative patterns common to the cohort. The loading coefficients differ across subjects, and evolve over time. At each timepoint/observation, they determine the contribution of each basis to the dynamic functional connectivity profile of the individual. Finally, the DTI Graph Laplacians re-weight the decomposition to focus on the functional connectivity between anatomically linked regions. The gray box denotes the deep networks part of our model. This network combines a Long Short Term Memory (LSTM) module with an Artificial Neural Network (ANN) to predict multiple behavioral scores. The LSTM models the temporal trends in the subject-specific loading coefficients giving rise to a hidden representation. The ANN then uses this representation to predict the corresponding behavioral outcomes.

\paragraph{\textbf{Dynamic Dictionary Learning for rs-fMRI data}}

We denote the set of time varying functional correlation matrices for individual $n$ by the set $\{\mathbf{\Gamma}^{t}_{n}\}^{T_{n}}_{t=1} \in \mathcal{R}^{P\times P}$. Here, $T_{n}$ denotes the number of sliding windows applied to the rs-fMRI scan, and $P$ is the number of ROIs in the parcellation scheme. As seen in Fig.~\ref{CCS} (green box), we model this information using a group average basis, and subject-specific temporal loadings. The dictionary $\mathbf{B} \in \mathcal{R}^{P \times K}$ is a concatenation of  $K$ elemental bases vectors $\mathbf{b}_{k} \in \mathcal{R}^{P \times 1}$, i.e. $\mathbf{B} := [\mathbf{b}_{1} \quad \mathbf{b}_{2} \quad ... \quad \mathbf{b}_{K}]$, where $K \ll P$. This basis captures representative brain states which each subject cycles through over the course of the scan. We further constrain the basis vectors to be orthogonal to each other. This constraint acts as an implicit regularizer, ensuring that the learned subnetworks are uncorrelated, yet explain the rs-fMRI data well. While the bases are shared across the cohort, the strength of their combination differs across individuals and varies over time. These loadings are denoted by the set $\{\mathbf{c}^{t}_{n}\}^{T_n}_{t=1}$ and combine the basis subnetworks uniquely to best explain each subject's functional connectivity. We introduce an explicit non-negativity constraint $\mathbf{c}^{t}_{nk}$ to ensure that the positive semi-definiteness of $\mathbf{\Gamma}^{t}_{n}$ is preserved. The complete rs-fMRI data representation takes the following form:
\begin{equation}
    \mathbf{\Gamma}^{t}_{n} \approx \sum_{k}{\mathbf{c}^{t}_{nk}\mathbf{b}_{k}\mathbf{b}_{k}^{T}} \ \ \  s.t.  \ \ \ \mathbf{c}_{nk}\geq 0, \ \ \mathbf{B}^{T}\mathbf{B} = \mathcal{I}_{K},
    \label{eqn:fMRI}
\end{equation}
where $\mathcal{I}_{K}$ is the $K \times K$ identity matrix. As seen in Eq.~(\ref{eqn:fMRI}), the subject-specific loading vector at time $t$, $\mathbf{c}^{t}_{n}: = [\mathbf{c}^{t}_{n1}\quad ... \quad \mathbf{c}^{t}_{nK}]^{T} \in \mathcal{R}^{K \times 1}$ models the heterogeneity in the cohort. Denoting $\textbf{diag}(\mathbf{c}^{t}_{n})$ as a diagonal matrix with the $K$ subject-specific coefficients on the diagonal and off-diagonal terms set to zero, Eq.~(\ref{eqn:fMRI}) can be re-written in the following matrix form: 
\begin{equation}
    \mathbf{\Gamma}^{t}_{n} \approx {\mathbf{B}\textbf{diag}({\mathbf{c}^{t}_{n}})\mathbf{B}^{T}}
    \ \ \  s.t.  \ \ \ \mathbf{c}^{t}_{nk}\geq 0, \ \ \mathbf{B}^{T}\mathbf{B} = \mathcal{I}_{K}
    \label{eqn:fMRI_mat}
\end{equation}
Finally, this matrix factorization serves to reduce the dimensionality of the rs-fMRI data, while simultaneously modeling group-level and subject-specific information.

\paragraph{\textbf{Structural Regularization from DTI data}}
{
Let $\mathbf{A}_{n} \in \mathcal{R}^{P \times P}$ be a binary adjacency matrix derived from the structural connectome of subject~$n$. For example, $\mathbf{A}_{n}$ can be constructed by thresholding the number of fibers estimated between two regions via tractography. Let $\mathcal{E}$ denote the set of edges in this graph. We compute the corresponding Normalized Graph Laplacian [Banerjee and Jost (2008)]  as $\mathbf{L}_{n} = \mathbf{V}_{n}^{-\frac{1}{2}}(\mathbf{V}_{n}-\mathbf{A}_{n})\mathbf{V}_{n}^{-\frac{1}{2}}$, where $\mathbf{V}_{n} = \mathbf{diag}(\mathbf{A}_{n}\mathbf{1})$ is the degree matrix and $\mathbf{1}$ is the vector of all ones. Intuitively, the Graph Laplacian is a discrete analog of the Laplace difference operator in Euclidean space. The Laplace difference operator has been used to characterize local properties of functions in Euclidean space (for example, to easily identify and characterize local optima). The Graph Laplacian generalizes this notion to discrete graphs and functions that are defined on graphs. Specifically, the Graph Laplacian has become a popular spatial regularizer in computer vision [\cite{pang2017graph}], genetics [\cite{feng2017pca}] and neuroimaging [\cite{atasoy2016human};  \cite{cuingnet2012spatial}]. This regularization implicitly assumes that there is a data signal associated with each node of the graph, and it encourages these signals to be similar for nodes of the graph that have an edge between them.}

\par {We use a matrix analog to Graph Laplacian regularization via the weighted Frobenius norm i.e. ${\vert\vert{.}\vert\vert}_{\mathbf{L}_{n}}$ [\cite{manton2003geometry,schnabel1983forcing}], which we use in place of the isotropic $\ell_{2}$ penalty in Eq.~(\ref{eqn:fMRI_mat}). In this case, the graph ``signal" corresponds to the vector (i.e., profile) of approximation errors given in Eq.~(\ref{eqn:fMRI_mat}) between the node in question and all other nodes in the graph. The underlying anatomical connectivity graph is defined by the DTI Graph Laplacian $\mathbf{L}_{n}$ for each patient. Mathematically, our dictionary learning loss takes the following form:
\begin{multline}
    \vert\vert{\mathbf{\Gamma}^{t}_{n} - {\mathbf{B}\textbf{diag}({\mathbf{c}^{t}_{n}})\mathbf{B}^{T}}}\vert\vert_{\mathbf{L}_{n}} \\ = \Tr{\Big[({\mathbf{\Gamma}^{t}_{n} - {\mathbf{B}\textbf{diag}({\mathbf{c}^{t}_{n}})\mathbf{B}^{T}}})\mathbf{L}_{n}({\mathbf{\Gamma}^{t}_{n} - {\mathbf{B}\textbf{diag}({\mathbf{c}^{t}_{n}})\mathbf{B}^{T}}})}\Big]
    \label{eqn:WFN}
\end{multline}
Here, $\Tr[{\mathbf{M}}]$ is the trace operator, which sums the diagonal elements of the argument matrix $\mathbf{M}$. For convenience, let $\mathbf{E}_{n}^{t} = {\mathbf{\Gamma}^{t}_{n} - {\mathbf{B}\textbf{diag}({\mathbf{c}^{t}_{n}})\mathbf{B}^{T}}}$ denote the element-wise approximation error of the the correlation matrix $\mathbf{\Gamma}_{n}^{t}$. Similarly, we define $\tilde{\mathbf{E}}^{t}_{n} = \mathbf{V}^{-\frac{1}{2}}_{n}\mathbf{E}^{t}_{n}$ as a weighted version of this error based on the degree matrix. As detailed in Appendix~A, Eq.~(\ref{eqn:WFN}) can be expanded as follows:
\begin{multline}
    \vert\vert{\mathbf{\Gamma}^{t}_{n} - {\mathbf{B}\textbf{diag}({\mathbf{c}^{t}_{n}})\mathbf{B}^{T}}}\vert\vert_{\mathbf{L}_{n}}  \\ = \sum_{(i,k)\in \mathcal{E}}{\vert\vert\tilde{\mathbf{E}}^{t}_{n}(i,:) 
    -\tilde{\mathbf{E}}^{t}_{n}(k,:)\vert\vert_{2}^2}
    \\
    = \sum_{(i,k) \in \mathcal{E}}\vert\vert{[\mathbf{V}_{n}(i,i)]^{-\frac{1}{2}} \mathbf{E}^{t}_{n}(i,:)- [\mathbf{V}_{n}}(k,k)]^{-\frac{1}{2}} \mathbf{E}^{t}_{n}(k,:)\vert\vert^2_{2}
    \label{eqn:err2}
\end{multline}
Notice that for terms where $(i,k) \not\in \mathcal{E}$, i.e. there is no anatomical connection between nodes $i$ and $k$, the corresponding error term in the summation drops out.
Said another way, this construction minimizes the sum of the square difference between the rs-fMRI reconstruction profiles ($\tilde{\mathbf{E}}^{t}_{n}(i,:)$ and $\tilde{\mathbf{E}}^{t}_{n}(k,:)$) between  nodes ($i$ and $k$) that are adjacent via the DTI graph. This effectively re-weights the rs-fMRI reconstruction profiles of anatomically connected nodes according to their relative degrees ($\mathbf{V}_{n}(i,i)$ and $\mathbf{V}_{n}(k,k)$) in the DTI graph pairwise. Thus, the functional connectivity at a particular node is directly influenced by its anatomical connections with other nodes in the graph. At a high level, this construction implicitly regularizes the rs-fMRI reconstruction loss according to the underlying anatomical connectivity prior.}

\par Finally, based on the formulation in Eq.~(\ref{eqn:WFN}), the final sr-DDL objective $\mathcal{D}(.)$ can be expressed as follows:
\begin{multline}
    \mathcal{D}(\mathbf{B},\{\mathbf{c}^{t}_{n}\};\{\mathbf{\Gamma}^{t}_{n}\},\mathbf{L}_{n}) \\ = \sum_{t}{\frac{1}{T_{n}}}{\vert\vert{\mathbf{\Gamma}^{t}_{n} - \mathbf{B}\mathbf{diag}(\mathbf{c}^{t}_{n})\mathbf{B}^{T}}\vert\vert}_{\mathbf{L}_{n}}
       \\ \ \  s.t.  \ \ \ \mathbf{c}^{t}_{nk}\geq 0, \ \ \mathbf{B}^{T}\mathbf{B} = \mathcal{I}_{K}
    \label{eqn:Dictionary Loss}
\end{multline}

\paragraph{\textbf{Deep Multiscore Prediction}}
As seen in the gray box in Fig.~\ref{CCS}, the subject-specific coefficients $\{\mathbf{c}^{t}_{n}\}$are input to an LSTM-ANN to predict the clinical scores, as parametrized by the weights $\mathbf{\Theta}$. The $M$ clinical scores for each individual are concatenated into a vector $\mathbf{y}_{n} := [\mathbf{y}_{n1}\quad ... \quad\mathbf{y}_{nM}]^{T} \in \mathcal{R}^{M \times 1}$. The LSTM models the temporal variations in the coefficients $\{\mathbf{c}^{t}_{n}\} $ to generate a hidden representation $\{\mathbf{h}^{t}_{n}\}^{T_n}_{t=1}$. From here, the Predictor ANN (P-ANN) generates a time varying estimates of the scores $\{\hat{\mathbf{y}}^{t}_{n}\}^{T_{n}}_{t=1} \in \mathcal{R}^{M \times 1}$. At the same time, the Attention ANN (A-ANN) generates $T_{n}$ scalars from the hidden representation. These are then softmax across time to obtain the attention weights: $\{a_{n}^{t}\}^{T_{n}}_{t=1}$. The final prediction is an attention-weighted average across the time estimates, which takes the following form:
\begin{equation}
{\hat{\mathbf{y}}}_{n} = \sum_{t}{\hat{\mathbf{y}}^{t}_{n} a_{n}^{t}}    
\end{equation}
Effectively, the attention weights determine which time points for each subject are most relevant for behavioral prediction. Additionally, they allow us to handle rs-fMRI scans of varying durations. Mathematically, we compute the multi-score prediction error $\mathcal{L}(.)$ using the Mean Squared Error (MSE) loss function as follows:
\begin{equation}
\mathcal{L}(\{\mathbf{c}^{t}_{n}\},\mathbf{y}_{n};\mathbf{\Theta})  =  {\vert\vert{\mathbf{\hat{y}}_{n}-\mathbf{{y}}_{n}}\vert\vert}^{2}_{F} = {\Bigg|\Bigg|{{\sum^{T_n}_{t=1}{\mathbf{\hat{y}}^{t}_{n} a^{t}_{n}}-\mathbf{{y}}_{n}}\Bigg|\Bigg|}^{2}_{F}}
\label{eqn:MSE}        
\end{equation}
At a high level, the deep network distills the temporal information to best predict each subject's clinical profile.

\par {We would like to highlight that our choice of the LSTM over a Recurrent Neural Network (RNN) allows us to track the temporal evolution of connectivity over longer horizons, while avoiding issues with convergence [\cite{chung2014empirical}]. Our two branched ANN in conjunction with the LSTM directly pools together time-varying estimates of clinical severity by focusing on the portions of the rs-fMRI scan most relevant to prediction. We notice that this construction naturally allows us to handle scans of varying length, while at same time obviating the need for additional sequence padding as would be required by a competing $1D$ CNN.}

\par {In Section~\ref{Optim}, we will develop a coupled optimization procedure to jointly estimate our unknowns $\{\mathbf{B},\{\mathbf{c}^{t}_{n}\},\mathbf{\Theta}\}$. We will show that our estimation procedure for the coefficients and neural network weights only relies on backpropagated gradients from the neural network loss and the parametric gradients from the dictionary learning. From the joint objective in Eq.~(\ref{eqn:JointObj}), we can see that the choice of neural network architecture does not directly affect the dictionary learning gradients. So long as we can backpropagate the deep network loss to the coefficients $\mathbf{c}^{t}_{n}$, we can effectively adopt our optimization strategy to handle an alternative architecture. Said another way, our coupled optimization procedure is agnostic to the specific neural network choice.}

{\paragraph{\textbf{Architectural Details}} Our proposed ANN architecture is highlighted in the white box to the bottom left of Fig.~\ref{CCS}. Our modeling choices carefully control for representational capacity and convergence of our coupled optimization procedure. Since the input to the network, i.e. the coefficient vector $\mathbf{c}^{t}_{n}$ is essentially low dimensional, we opt for a two layered LSTM with the hidden layer width as $40$. Both the P-ANN and the A-ANN are fully connected neural networks with two hidden layers of width $40$. Since the A-ANN outputs a scalar, the width of its output layer is one, while that of the P-ANN is of size $M$, i.e. the number of behavioral scores. We use a Rectified Linear Unit (ReLU) as the activation function for each hidden layer, as we found that this choice is robust to issues with vanishing gradients and saturation that commonly confound the training of deep neural networks [\cite{glorot2011deep}].}

\paragraph{\textbf{Joint Objective for Multimodal Integration}}
We combine the complementary viewpoints in Eq.~(\ref{eqn:Dictionary Loss}) and Eq.~(\ref{eqn:MSE}) into a single joint objective below:
\begin{multline}
\mathcal{J}(\mathbf{B},\{\mathbf{c}^{t}_{n}\},\mathbf{\Theta};\{\mathbf{\Gamma}^{t}_{n}\},\mathbf{L}_{n}, \{\mathbf{y}_{n}\}) \\ = \underbrace{\sum_{n} \mathcal{D}(\mathbf{B},\{\mathbf{c}^{t}_{n}\};\{\mathbf{\Gamma}^{t}_{n}\},\mathbf{L}_{n})}_{\textbf{sr-DDL loss}} + \lambda \underbrace{\sum_{n}\mathcal{L}(\mathbf{\Theta},\{\mathbf{c}^{t}_{n}\};\mathbf{y}_{n})}_{\textbf{deep network loss}} \ \ \ \ 
\\ = \sum_{n}\sum_{t}{\frac{1}{T_{n}}}{\vert\vert{\mathbf{\Gamma}^{t}_{n} - \mathbf{B}\mathbf{diag}(\mathbf{c}^{t}_{n})\mathbf{B}^{T}}\vert\vert}_{\mathbf{L}_{n}} \\ + \lambda \sum_{n}{\mathcal{L}(\mathbf{\Theta},\{\mathbf{c}^{t}_{n}\};\mathbf{y}_{n})}
\ \   s.t.  \ \ \ \mathbf{c}^{t}_{nk}\geq 0, \ \ \mathbf{B}^{T}\mathbf{B} = \mathcal{I}_{K}
\label{eqn:JointObj}
\end{multline}
Here, $\lambda$ is a hyperparameter than balances the tradeoff between the representation loss $\mathcal{D}(.)$ and the prediction loss $\mathcal{L}(.)$. $\{\mathbf{B},\{\mathbf{c}_{n}^{t}\},\mathbf{\Theta}\}$ are the variables to optimize.

\begin{figure*}[t!]
   \centering
   \includegraphics[width=\dimexpr \textwidth-2\fboxsep-2\fboxrule\relax]{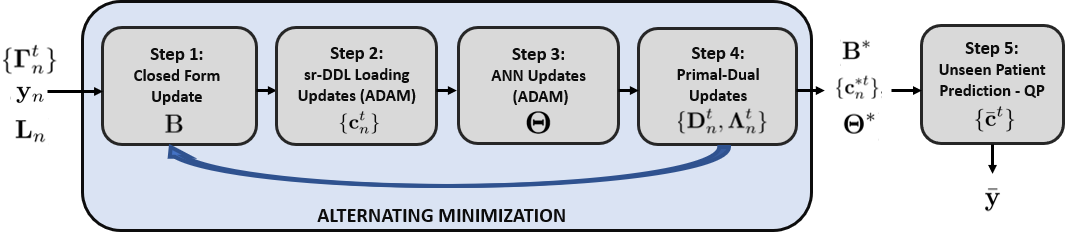}
   \caption{{Alternating minimization strategy for joint optimization of Eq.~(\ref{eqn:const}})  \label{fig:AltMin}}
\end{figure*}

\subsection{{Coupled Optimization Strategy}}
\label{Optim}
\par We employ the alternating minimization technique in order to infer the set of hidden variables $\{\mathbf{B},\{\mathbf{c}^{t}_{n}\},\mathbf{\Theta}\}$. Namely, we optimize Eq.~(\ref{eqn:JointObj}) for each output variable, while holding the other unknowns constant.

\par We utilize the fact that there is a closed-form Procrustes solution for quadratic objectives of the form ${\vert\vert{\mathbf{M}-\mathbf{B}}\vert\vert}^{2}_{F}$~[\cite{everson1998orthogonal}]. However, Eq.~(\ref{eqn:JointObj}) is bi-quadratic in $\mathbf{B}$, so it cannot be directly applied. Therefore, we adopt the strategy in [\cite{d2019coupled,d2019integrating,d2020joint}] of introducing $\sum_{n}{T_{n}}$ constraints of the form $\mathbf{D}^{t}_{n} = \mathbf{B}\mathbf{diag}(\mathbf{c}^{t}_{n})$. These constraints are enforced via the Augmented Lagrangian algorithm with corresponding constraint variables $\{\mathbf{\Lambda}^{t}_{n}\}$. Thus, our objective from Eq.~(\ref{eqn:JointObj}) now becomes:
\begin{multline}
\mathcal{J}_{c} = \sum_{n,t}{\frac{1}{T_{n}}}{\vert\vert{\mathbf{\Gamma}^{t}_{n} - \mathbf{D}^{t}_{n}\mathbf{B}^{T}}\vert\vert}_{\mathbf{L}_{n}} + \lambda \sum_{n}\mathcal{L}(\mathbf{\Theta},\{\mathbf{c}^{t}_{n}\};\mathbf{y}_{n}) \\ 
+ \sum_{n,t}{\frac{\gamma}{T_{n}}\Big[{\Tr{\left[{(\mathbf{\Lambda}^{t}_{n})^{T}({\mathbf{D}^{t}_{n}-\mathbf{B}\mathbf{diag}(\mathbf{c}^{t}_{n})})}\right]}}\Big]} \\ + \sum_{n.t}{\frac{\gamma}{T_{n}}}{\Big[{{\frac{1}{2}}~{\vert\vert{\mathbf{D}^{t}_{n}-\mathbf{B}\mathbf{diag}(\mathbf{c}^{t}_{n})}\vert\vert}_{F}^{2}}}\Big] \\
\ \  s.t.  \ \ \mathbf{c}^{t}_{nk} \geq 0, \mathbf{B}^{T}\mathbf{B} = \mathcal{I}_{K} 
\label{eqn:const}
\end{multline}

The Frobenius norm terms ${\vert\vert{\mathbf{D}^{t}_{n}-\mathbf{B}\mathbf{diag}(\mathbf{c}^{t}_{n})}\vert\vert}_{F}^{2}$ regularize the trace constraints during the optimization. Observe that Eq.~(\ref{eqn:const}) is convex in the set $\{\mathbf{D}^{t}_{n}\}$, which allows us to optimize this variable via standard procedures. The constraint parameter is fixed at $\gamma=20$, based on the guidelines in the literature [\cite{nocedal2006numerical}].

\par {
Fig.~\ref{fig:AltMin} depicts our alternating minimization strategy. We describe each individual block in detail below. We refer the interested reader to Appendix~B, which systematically delineates the supporting calculations from this section:
\paragraph{\textbf{{Step~{1}: Closed form solution for $\mathbf{B}$}}} Notice that Eq.~(\ref{eqn:const}) reduces to the following quadratic form in $\mathbf{B}$:
\begin{equation}
\mathbf{B}^{*} = \argmin_{\mathbf{B} : \ \mathbf{B}^{T}\mathbf{B}=\mathcal{I}_{K}}{{\vert\vert{\mathbf{M}-\mathbf{B}}\vert\vert}^{2}_{F}}
\end{equation}
Given the singular value decomposition $\mathbf{M} = \mathbf{U}\mathbf{S}\mathbf{V}^{T}$, we have the following closed form solution :
\begin{equation*}
    \mathbf{B}^{*} = \mathbf{U}\mathbf{V}^{T}
\end{equation*}
where $\mathbf{M}$ is computed as follows:
\begin{multline}
    \mathbf{M} = \sum_{n}{\frac{1}{T_{n}}}\sum_{t}{(\mathbf{\Gamma}^{t}_{n}\mathbf{L}_{n}+ \mathbf{L}_{n}\mathbf{\Gamma}^{t}_{n})\mathbf{D}^{t}_{n}} +  \\ \sum_{n}{\frac{1}{T_{n}}}\Big[\sum_{t}{\frac{\gamma}{2}\mathbf{D}^{t}_{n}\mathbf{diag}(\mathbf{c}^{t}_{n})+ \gamma\mathbf{\Lambda}^{t}_{n}\mathbf{diag}(\mathbf{c}^{t}_{n})}\Big] 
\end{multline} 
Essentially, $\mathbf{B}$ spans the anatomically weighted space of subject-specific dynamic correlation matrices.}

{\paragraph{\textbf{Step~{2}: Updating the sr-DDL loadings  $\{\mathbf{c}_{n}^{t}\}$}} The objective $\mathcal{J}_{c}$ in Eq.~(\ref{eqn:const}) decouples across subjects. Additionally, we can also incorporate the non-negativity constraint $\mathbf{c}^{t}_{nk} \geq 0$ by passing an intermediate vector $\hat{\mathbf{c}}^{t}_{n}$ through a ReLU. The ReLU pre-filtering allows us to optimize an unconstrained version of Eq.~(\ref{eqn:const}), which can be done via the stochastic ADAM algorithm [\cite{kingma2015adam}]. In essence, this optimization couples the parametric gradient from the augmented Lagrangians with the backpropagated gradient from the deep network (defined by fixed $\mathbf{\Theta}$). After convergence, the thresholded loadings ${\mathbf{c}}^{t}_{n} = ReLU(\hat{\mathbf{c}}^{t}_{n})$ are used in subsequent steps.}

{\paragraph{\textbf{Step~{3}: Updating the Deep Network weights-$\mathbf{\Theta}$}}  We backpropagate the loss $\mathcal{L(\cdot)}$ to solve for the unknowns $\mathbf{\Theta}$. Notice that by dropping the contributions of the unknown value of $\mathbf{y}_{nm}$ to the network loss during backpropagation using the ADAM [\cite{kingma2015adam}] algorithm, we can handle missing clinical data as well. }

{\paragraph{\textbf{Step~{4}: Updating the Constraint Variables $\{\mathbf{D}^{t}_{n},\mathbf{\Lambda}^{t}_{n}\}$}}We perform parallel primal-dual updates for the constraint pairs $\{\mathbf{D}^{t}_{n},\mathbf{\Lambda}^{t}_{n}\}$. Here, we cycle through the closed form update for $\mathbf{D}_{n}^{t}$ and gradient ascent for $\mathbf{\Lambda}^{t}_{n}$ until convergence.}

{\paragraph{\textbf{Step~{5}: Prediction on Unseen Data}}
In our cross-validated setting, we need to compute the sr-DDL loadings $\{\mathbf{\Bar{c}}^{t}\}_{t=1}^{\bar{T}}$ for a new patient based on the training $\mathbf{B}^{*}$. Since we do not know the score $\mathbf{\bar{y}}$ for this patient, we remove the contribution $\mathcal{L}(\cdot)$ from Eq.~(\ref{eqn:JointObj}) and assume the constraints $\bar{\mathbf{D}}^{t} = \mathbf{B^{*}}\mathbf{diag}(\bar{\mathbf{c}}^{t})$ hold with equality, thus removing the Lagrangian terms. Essentially, the optimization for $\{\mathbf{\Bar{c}}^{t}\}$ reduces to decoupled quadratic programming (QP)~objectives~$\mathcal{Q}_{t}$~across~time: \begin{eqnarray*}
\Bar{\mathbf{c}}^{*t} =\argmin_{\bar{\mathbf{c}}^{t}}{\frac{1}{2}}~{(\mathbf{\Bar{c}}^{t})^{T}\mathbf{\Bar{H}}\mathbf{\Bar{c}}^{t}} + \mathbf{\Bar{f}}^{T}\mathbf{\Bar{c}}^{t} \ \ s.t.\ \  \mathbf{\Bar{A}}\mathbf{\Bar{c}}^{t} \leq \mathbf{\Bar{b}} \notag
\\ 
\mathbf{\Bar{H}} = 2(\mathbf{B^{*}}^{T}\bar{\mathbf{L}}\mathbf{B^{*}}); \ \ \ \ \ \ \ \ \ \ \ \ \ \ \nonumber \\ \ \ \ \ \ \ \ \ \mathbf{\Bar{f}} = -[\mathcal{I}_{K} \circ (\mathbf{B^{*}}^{T}(\bar{\mathbf{\Gamma}^{t}}\bar{\mathbf{L}}+ \bar{\mathbf{L}}\bar{\mathbf{\Gamma}^{t}})\mathbf{B^{*}})]\mathbf{1};   \ \ \ \ \ \ \ \ \ \ \ \ \ \   \\ \mathbf{\Bar{A}} = -\mathcal{I}_{K}  
 \   \mathbf{\Bar{b}} = \mathbf{0}  \ \ \ \ \ \ \ \ \ \ \ \ \ \ \
\label{eqn:QP}
\end{eqnarray*}
Where, $\circ$ denotes the Hadamard product. Finally, we  estimate $\bar{\mathbf{y}}$ via a forward pass through the LSTM-ANN.}

\par {Overall, our alternating minimization training procedure explicitly couples the Dictionary Learning (sr-DDL) and Deep Network (LSTM-ANN) blocks within the optimization. In contrast, the setup at test time consists of two steps, namely the coefficient update followed by a forward pass through the LSTM-ANN. We will demonstrate via our experiments (i.e. Section~\ref{Expt}) that the coupled training is key to generalization. Finally, we discuss the effect of this difference between the training and testing procedures further in Section~\ref{Gen}}

\subsubsection{Implementation Details}
\label{Implementation}

\begin{figure}[b!]
  \centering
    \includegraphics[width=0.46\textwidth]{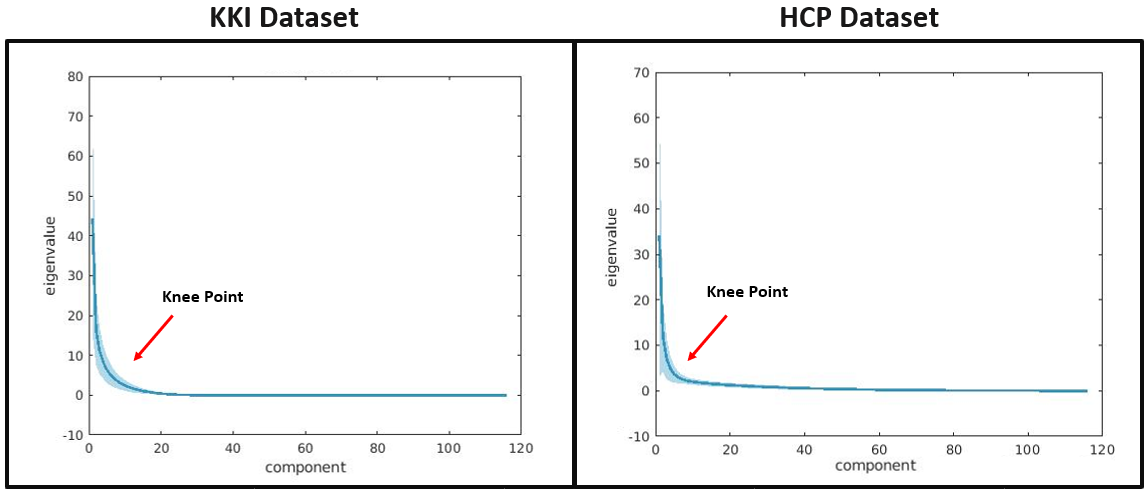}
    \caption{Scree Plot of the correlation matrices to corroborate the selected values for $K$. \textbf{(L)} KKI Dataset \textbf{(R)} HCP Dataset. The thick line denotes the mean eigenvalue, while the shaded area indicates the standard deviation across subjects and time points.}
    \label{fig:ScreePlot}
\end{figure}

\paragraph{\textbf{Parameter Settings:}} 
{In order to fix the hyperparameters for our model and the baselines, we make use of a second subset of $130$ individuals from the HCP database (hereby referred to as HCP-$2$). Note that these individuals have no overlap with those used characterize the performance in Section~\ref{Expt} to avoid biasing the results. First, we set aside $30$ of these patients as a validation set to determine appropriate learning rates for our method and baselines. Recall that our deep-generative hybrid has two free parameters: namely the penalty $\lambda$, which controls the tradeoff between data representation and clinical prediction, and $K$, the number of networks. For our experiments, we chose $K= 15$ for both datasets  based  on the knee point of the eigenspectrum of the correlation matrices $\{\mathbf{\Gamma}^{t}_{n}\}$ (See Fig.~\ref{fig:ScreePlot}). Based on the results of a $5$ fold cross validation and grid search on HCP-$2$, we fix $\lambda=2.5$. We will further discuss the robustness to $\lambda$ in Section~\ref{Sensitivity}. Along similar lines, our}  {Section~\ref{AppdD}} {includes a discussion on emerging subnetwork patterns in $\mathbf{B}$ upon varying the model order, i.e. $K$.} 
\par {Additionally, our sliding window protocol is defined by two parameters, namely the window length and stride. Although these are not hyperparameters for the sr-DDL per se, they affect the predictive performance by controlling the information overlap between successive dynamic rs-fMRI correlation matrices. Again, these are set based on the cross validation performance on HCP-2. We will further discuss the robustness to these parameters in Section~\ref{Sensitivity}.}

\paragraph{\textbf{Initialization:}} Our coupled optimization strategy requires us to initialize the basis $\mathbf{B}$, coefficients $\{\mathbf{c}^{t}_{n}\}$, the deep network weights $\mathbf{\Theta}$ and the constraint variable pairs $\{\mathbf{D}^{t}_{n},\mathbf{\Lambda}^{t}_{n}\}$. We randomly initialize the deep network weights at the first main iteration. We employ a soft-initialization for $\{\mathbf{B},\{\mathbf{c}^{t}_{n}\}\}$ by solving the dictionary objective in Eq.~(\ref{eqn:Dictionary Loss}) without the LSTM-ANN loss terms for $20$ iterations. We then initialize $\mathbf{D}^{t}_{n} =\mathbf{B}\mathbf{diag}(\mathbf{c}^{t}_{n})$ and $\mathbf{\Lambda}^{t}_{n} = \mathbf{0}$ which lie in the feasible set for our constraints. We empirically observed that this soft initialization helps stabilize the optimization to provide improved predictive performance in fewer main iterations when compared with a completely random initialization.

Finally, the meta-data and code used in this study are available on a public repository hosted on Github \footnote{\href{https://github.com/Niharika-SD/Deep-sr-DDL}{https://github.com/Niharika-SD/Deep-sr-DDL}}.
\begin{figure*}[t!]
   \centering
   \fbox{\includegraphics[width= \textwidth- 2cm]{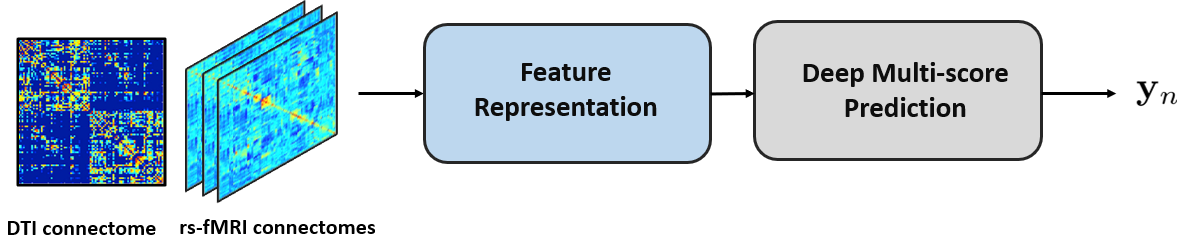}}
   \caption{A typical two stage baseline. We input the dynamic correlation matrices and DTI connectomes to Stage~$1$, which performs Feature Extraction. This step could be a technique from machine learning, graph theory or a statistical measure. Stage~$2$ is a deep network that predicts the clinical scores }
   \label{fig:BS}
\end{figure*}

\subsection{Baseline Comparison Techniques}
\label{baselines}
We evaluate the performance of our framework against three different classes of baselines, each highlighting the benefit of specific modeling choices made by our method. 

\par Our first baseline class is a two stage configuration as illustrated in Fig.~\ref{fig:BS} that combines feature extraction on the dynamic rs-fMRI and DTI data, with a deep learning predictor. These feature engineering techniques are drawn from a set of well established statistical (Independent Component Analysis in Subsection~\ref{ICA}) and graph theoretic techniques (Betweenness Centrality in Subsection~\ref{GT}), known to provide rich feature representations. The learned features are then input to the same deep LSTM-ANN network used by our method. This network is trained separately to predict the clinical outcomes. Note that these baselines incorporate multimodal and dynamic information, but do not directly operate on the network structure of the connectomes. Our second baseline class omits the two step approach in lieu of an end-to-end convolutional neural network based on the work of  [\cite{kawahara2017brainnetcnn}]. We train this model on the static rs-fMRI and DTI connectomes in tandem to predict the clinical scores. This baseline operates directly on the correlation and connectivity matrices, but ignores the dynamic evolution of functional connectivity. Next, we present the comparison of our deep sr-DDL by omitting the structural regularization. This helps us evaluate the benefit provided by the multimodal integration of DTI and rs-fMRI data. Our final baseline highlights the benefit of our joint optimization procedure. In this experiment, we decouple the optimization of the dynamic matrix factorization and deep network in Fig.~\ref{CCS} similar to the two stage pipelines.

\subsubsection{Graph Theoretic Feature Selection:}
\label{GT}

Notice that the subject-specific correlation rs-fMRI matrices $\{\mathbf{\Gamma}^{t}_{n}\}$ and the corresponding binary DTI adjacency matrices $\mathbf{A}_{n}$ indicate time-varying functional and anatomical connectivity between the ROIs respectively. Therefore, we multiply the two to generate the time-varying multimodal graphs whose nodes are the brain ROIs and edges are defined by the temporal connectivity between these ROIs. We denote the corresponding adjacency matrices for these graphs by $\{\mathbf{\Psi}_{n}^{t} = \mathbf{A}_{n} \circ \mathbf{\Gamma}^{t}_{n} \in \mathcal{R}^{P \times P}\}$, where we threshold each $\mathbf{\Psi}^{t}_{n}$ to remove negative values. Each element $[\mathbf{\Psi}_{n}^{t}]_{ij}$ gives the strength of association between two communicating sub-regions $i$ and $j$ in individual $n$ at time $t$. We summarize the topology of these graphs via \textbf{Betweenness Centrality ($C_{B}$)} to obtain a time-varying estimate of brain connectivity for each ROI [\cite{sporns2004organization,bassett2006small}].  $\mathbf{C}_{B}(v)$ for region $v$ is calculated as:
\begin{equation}
\mathbf{C}^{t}_{B}(v) = \sum_{s \neq v \neq u\in V} {\frac{\mathbf{\sigma}^{t}_{su}(v)}{\mathbf{\sigma}^{t}_{su}}}
\end{equation}
$\mathbf{\sigma}^{t}_{su}$ is the total number of shortest paths from node $s$ to node $u$ at time $t$, and $\mathbf{\sigma}^{t}_{su}(v)$ is the number of those paths that pass through $v$. This measure quantifies the number of times a node acts as a bridge along the shortest path between two other nodes and has found wide usage in characterizing small-worlded networks in brain connectivity [\cite{sporns2004organization}]. We effectively reduce the dimensionality of the connectivity features. Again, the collection of features $\{\mathbf{C}^{t}_{B}\}$ are used to train an LSTM-ANN predictor from Fig.~\ref{CCS} with two hidden layers having width $200$ due to the higher input feature dimensionality.

\subsubsection{ICA Feature Selection}
\label{ICA}
This baseline employs \textbf{Independent Component Analysis (ICA)} combined an the LSTM-ANN predictor. ICA is a statistical technique that extracts representative spatial patterns from the rs-fMRI time series. It has now become ubiquitous in fMRI analysis for its ability to identify group level differences as well as model individual-specific connectivity signatures. Essentially, ICA decomposes multivariate signals into `independent' non-Gaussian components based on the data statistics.

\begin{figure*}[t!]
   \centering
   \fbox{\includegraphics[width= \textwidth- 2cm]{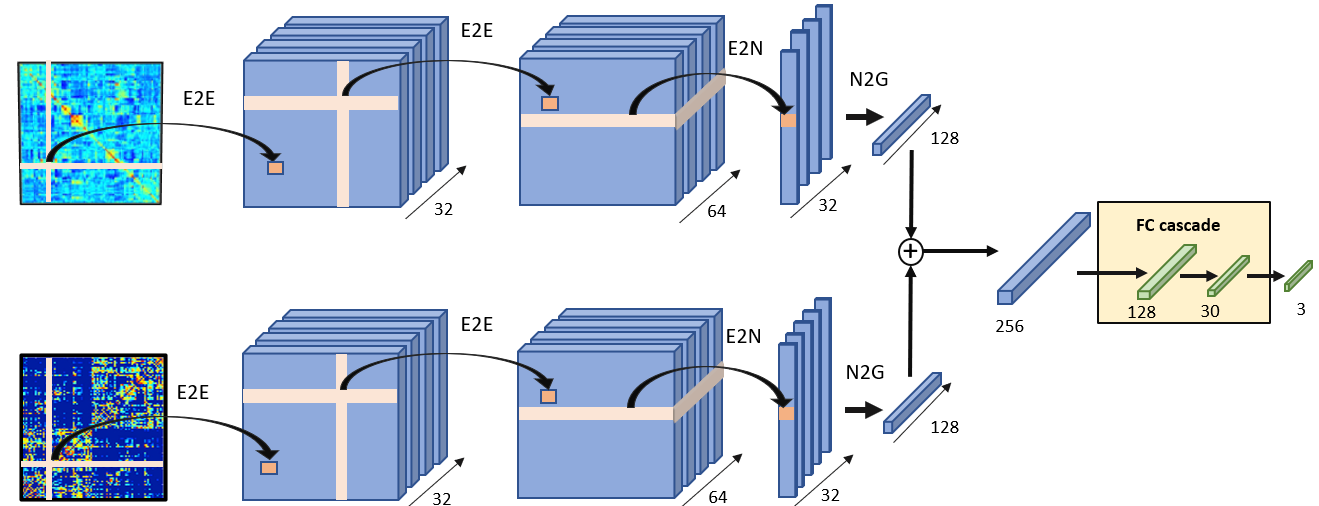}}
   \caption{The BrainNet CNN baseline [\cite{kawahara2017brainnetcnn}] for severity prediction from multimodal data}
   \label{fig:Bncnn}
\end{figure*}

\par This algorithm can be extended to the multi-subject analysis setting via Group ICA (G-ICA). Specifically, we extract independent spatial patterns common across patients, by combining the contribution of the individual time courses. For this baseline, we first perform G-ICA using the GIFT toolbox [\cite{calhoun2009review}], and derive independent spatial maps for each subject from their raw rs-fMRI scans. We then compute the average time courses for each spatial map considering the constituent voxels. This provides us with a feature representation of reduced dimension equal to the number of specified maps ($d<<L$) for each individual. For our experiments, we extract $15$ ICA components. These time courses are input into the LSTM-ANN network in Fig.~\ref{CCS} with two hidden layers of width $40$ to predict the clinical outcomes. 

\subsubsection{BrainNet Convolutional Neural Network}

The BrainNet CNN [\cite{kawahara2017brainnetcnn}] relies on specialized fully convolutional layers for feature extraction, and was originally used to predict cognitive and motor outcomes from DTI connectomes. Fig.~\ref{fig:Bncnn} provides a pictorial overview of the original architecture adapted for clinical outcome prediction from multimodal data. Each branch of the network accepts as input a $P\times P$ connectome, to which it applies a cascade of two edge-edge (E-E) convolutional operations. This E-E operation combines individual convolutions acting on the row and column to which the input element belongs. It is followed by a series of edge-node (E-N) blocks that reduce the dimensionality of the intermediate outputs, followed by a node-graph (N-G) operation for pooling. Finally, the output clinical scores are predicted via a fully connected artificial neural network for regression.

\par We feed the rs-fMRI static connectomes ($\hat{\mathbf{\Gamma}}_{n}$) and DTI Laplacians $\mathbf{L}_{n}$ into two disjoint fully convolutional branches with the architecture described above. We integrate the learned features via concatenation and input them into the fully connected layers described in Fig.~\ref{fig:Bncnn}, but with the number of outputs equal to the dimensionality of the clinical severity vector $\mathbf{y}_{n}$. We set the learning rate, momentum and weight decay parameters according to the guidelines in [\cite{kawahara2017brainnetcnn}].

\subsubsection{Deep sr-DDL without DTI regularization}
In this baseline, we examine the effect of excluding the structural regularization provided by the DTI data from the joint objective in Eq.~(\ref{eqn:JointObj}). The resulting objective function takes the following form:
\begin{multline}
    \mathcal{J}_{w}(\mathbf{B},\{\mathbf{c}^{t}_{n}\},\mathbf{\Theta};\{\mathbf{\Gamma}^{t}_{n}\}, \{\mathbf{y}_{n}\}) \\= \sum_{n}\sum_{t}{\frac{1}{T_{n}}}{\vert\vert{\mathbf{\Gamma}^{t}_{n} - \mathbf{B}\mathbf{diag}(\mathbf{c}^{t}_{n})\mathbf{B}^{T}}\vert\vert}^{2}_{F} \\ + \lambda \sum_{n}{\mathcal{L}(\mathbf{\Theta},\{\mathbf{c}^{t}_{n}\};\mathbf{y}_{n})}
\ \   s.t.  \ \ \ \mathbf{c}^{t}_{nk}\geq 0, \ \ \mathbf{B}^{T}\mathbf{B} = \mathcal{I}_{K}.
\label{eqn:no_DTI}
\end{multline}
Notice that amounts to replacing the Weighted Frobenius Norm formulation by a regular $\ell_{2}$ penalty. This allows us to adopt the alternating minimization procedure in Section~\ref{Optim} to optimize Eq.~(\ref{eqn:no_DTI}) with a few minor modifications. Specifically, instead of $T_{n}$ constraints per subject, we use a single constraint of the form $\mathbf{D}=\mathbf{B}$, enforced via a single Augmented Lagrangian $\mathbf{\Lambda}$. This effectively ensures that the new objective has a quadratic form in $\mathbf{B}$, along with a closed form update for $\mathbf{D}$. As before, we cycle through four individual steps, namely:
\begin{itemize}
    \item Closed form Procrustes solution for the basis $\mathbf{B}$
    \item Updating the temporal loadings $\{\mathbf{c}^{t}_{n}\}$ (ADAM) 
    \item Updating the Neural Network Parameters $\mathbf{\Theta}$ (ADAM)
    \item Augmented Lagrangian updates for the constraint variables $\{\mathbf{D},\mathbf{\Lambda}\}$
\end{itemize}
Similar to the Deep sr-DDL, we use $K=15$ networks as inputs to the LSTM-ANN network with two hidden layers of width $40$ to predict the clinical outcomes.

\subsubsection{Decoupled Deep sr-DDL}
Our final baseline examines the efficacy of our coupled optimization procedure in Section~\ref{Optim} with regards to generalization onto unseen subjects. Here, we first run the feature extraction using the sr-DDL optimization to extract the basis $\mathbf{B}$ and temporal loadings $\{\mathbf{c}^{t}_{n}\}$. We then use the $\{\mathbf{c}^{t}_{n}\}$ as inputs to train the LSTM-ANN network in Fig.~\ref{CCS} to predict the scores $\mathbf{y}_{n}$. This is akin to the two-stage baselines delineated in Fig.~\ref{fig:BS}.
\par Again, we use $K=15$ networks with an a two layered LSTM-ANN having hidden layer width $40$
\section{Experimental Results:}

\subsection{Validation on Synthetic Data}
{As a sanity check, we first validate our optimization in Section~$2.2$ on synthetic data generated from the equivalent generative process. This experiment allows us to assess the behavior of our algorithm under various noise scenarios. Specifically, we evaluate the robustness of our estimation procedure under varying levels of noise in the correlation matrices and the scores, and under increasing deviations from orthogonality in our generating basis. Our simulations indicate that the optimization procedure is robust in the noise regime ($0.01-0.2$) estimated from the real-world rs-fMRI data. In addition, these experiments help us identify the stable parameter settings ($\lambda = 1-10$) which guide our real world experiments. We refer the interested reader to the Supplementary Results for the details from this section.}

\subsection{Real-World Experiments: Population Studies of Connectomics and Behavior}
\label{Expt}
We evaluate our deep-generative hybrid on two separate cohorts. {The first dataset is a cohort of $150$ healthy individuals from the Human Connectome Project (HCP) database [\cite{van2013wu}] having both the rs-fMRI and DTI scans. We refer to this as the HCP dataset.} Cognitive outcomes such as fluid intelligence are believed to be closely connected to structural (SC) and function connectivity (FC) in the human brain [\cite{zimmermann2018unique}]. Thus, jointly modeling multimodal neuroimaging and cognitive data helps exploit this fundamental interweave and uncover the neural underpinnings of cognition. {Finally, we chose to focus on a modest sized dataset ($N=150$) to demonstrate that our framework is suitable for clinical rs-fMRI applications, many of which have limited sample sizes.}

\par Our second dataset consists of $57$ children with high functioning Autism Spectrum Disorder (ASD) acquired at the Kennedy Krieger Institute in Baltimore, USA. Henceforth, we refer to this as the KKI dataset. The age of the subjects from this cohort is $10.06 \pm 1.26$ with an IQ of $110\pm14.03$. Social and communicative deficits in ASD are believed to arise from aberrant interactions between regions of the brain that are linked by structural and functional connectivity [\cite{rudie2013altered}]. Thus, identifying these patterns plays a crucial role in illuminating the etiological basis of the disorder. 

\paragraph{\textbf{Neuroimaging Data}}  
As described in [\cite{van2013wu}], the HCP S1200 dataset was acquired on a Siemens $3$T scanner (TR/TE$=0.72ms/0.33ms$, spatial resolution $=2 \times 2 \times 2 $mm). The rs-fMRI scans were processed according to the standard pre-processing pipeline described in [\cite{smith2013resting}], which includes additional processing to account for confounds due to motion and physiological noise. We opted to use a $15$ minute interval (typical of clinical rs-fMRI studies of neurodevelopmental disorders) from the second scan of each subject’s first visit for our analysis. 

\par The DTI data from the HCP dataset was processed using the standard Neurodata MR Graphs package (ndmg) [\cite{kiar2016ndmg}]. This consists of co-registration to anatomical space via FSL [\cite{jenkinson2012fsl}], followed by tensor estimation in the MNI space and probabilistic tractography to compute the fibre tracking streamlines.

\par For the KKI dataset, rs-fMRI acquisition was performed on a Phillips $3T$ Achieva scanner with a single shot, partially parallel gradient-recalled EPI sequence with TR/TE = $2500/30$ms, flip angle $\ang{70}$, res $=3.05 \times 3.15 \times 3$mm, having $128$ or $156$ time samples. The children were instructed to relax with eyes open and focus on a central cross-hair while remaining still. We used an in-house pre-processing pipeline pre-validated across several studies [\cite{nebel2016intrinsic,venkataraman2017unified,d2020joint}]. This consists of slice time correction, rigid body realignment, and normalization to the EPI version of the MNI template using SPM [\cite{penny2011statistical}], followed by temporal detrending of the time courses to remove gradual trends in the data. A CompCorr50 [\cite{muschelli2014reduction, ciric2018mitigating}] strategy was used to estimate and remove spatially coherent noise from the white matter and ventricles, along with the linearly detrended versions of the six rigid body realignment parameters and their first derivatives, followed by spatial smoothing using a $6$mm FWHM Gaussian kernel and temporal smoothing via a  band pass filter ($0.01-0.1$Hz). Lastly, the data was despiked using the AFNI package [\cite{cox1996afni}].

\par The DTI acquisition for the KKI dataset was collected on a 3T Philips scanner (EPI, SENSE factor$ = 2.5$, TR$ = 6.356$s, TE$ = 75ms$, res = $0.8\times 0.8 \times 2.2$mm, and FOV$= 212$). We collected two identical runs, each with a single b0 and $32$ non-collinear gradient directions at $b= 700s/mm^{2}$. The data was pre-processed using the standard FDT [\cite{jenkinson2012fsl}] pipeline in FSL consisting of susceptibility distortion correction, followed by corrections for eddy currents, motion and outliers. From here, tensor model fitting was performed to generate the transformation matrices and extract atlas based metrics. We used the BEDPOSTx tool in FSL [\cite{behrens2007probabilistic}] to perform a bayesian estimation of the diffusion parameters at each voxel, followed by tractography using PROBTRACKx [\cite{behrens2007probabilistic}].

\par Our experiments rely on the Automatic Anatomical Labelling (AAL) atlas [\cite{tzourio2002automated}] parcellation for the rs-fMRI and DTI data. AAL consists of $116$ cortical, subcortical and cerebellar regions. {We employ a sliding window protocol as shown in Fig.~\ref{Ex_Dyn} using the parameters learned in Section~\ref{Implementation}}. Due to the different TR, we set the sliding window parameters to window length~$=156$ and stride~$=17$ for the HCP dataset, and  window length~$=45$ and stride~$=5$ for the KKI dataset to extract dynamic correlation matrices from the $116$ average time courses. We discuss the sensitivity to this choice in Section~\ref{Sensitivity}. Thus, for each individual, we have correlation matrices of size $116 \times 116$ based on the Pearson's Correlation Coefficient between the average regional time-series. Empirically, we observed a consistent noise component with nearly unchanging contribution from all brain regions and low predictive power for both datasets. Therefore, we subtracted out the first eigenvector contribution from each of the correlation matrices and used the residuals as the inputs $\{\mathbf{\Gamma}_{n}\}$ to the algorithm and the baselines. 

\par Each DTI connectivity matrix $\mathbf{A}_{n}$ is binary, where $[\mathbf{A}_{n}]_{ij} = 1$ corresponds to the presence of at least one tract between the regions $i$ and $j$, $116$ in total for AAL. For the KKI dataset, we impute the DTI connectivity for the $11$ individual, who do not have DTI based on the training data in each cross validation fold.

\paragraph{\textbf{Behavioral Data}} For the HCP database, we examine the Cognitive Fluid Intelligence Score (CFIS) described in [\cite{duncan2005frontal, bilker2012development}], adjusted for age. This is scored based on a battery of tests measuring cognitive reasoning, considered a nonverbal estimate of fluid intelligence in subjects. The dynamic range for the score is $70-150$, with higher scores indicating better cognitive abilities.

\par We analyzed three independent measures of clinical severity for the KKI dataset. These include:
\begin{itemize}
\setlength\itemsep{0.1em}
    \item[1] Autism Diagnostic Observation Schedule, Version $2$ (ADOS-2) total raw score
    \item[2] Social Responsiveness Scale (SRS) total raw score
    \item[3] Praxis total percent correct score
\end{itemize} 

\par The ADOS consists of several sub-scores which quantify the social-communicative deficits in individuals along with the restrictive/repetitive behaviors [\cite{lord2000autism}]. The test evaluates the child against a set of guidelines and is administered by a trained clinician. We compute the total score by adding the individual sub-scores. The dynamic range for ADOS is between $0-30$, with higher score indicating greater impairment.

\par The SRS scale quantifies the level of social responsiveness of a subject [\cite{bolte2008assessing}]. Typically, these attributes are scored by parent/care-giver or teacher who completes a standardized questionnaire that assess various aspects of the child's behavior. Consequently, SRS reporting tends to be more variable across subjects, as compared to ADOS, since the responses are heavily biased by the parent/teacher attitudes. The SRS dynamic range is between $70-200$ for ASD subjects, with higher values corresponding to higher severity in terms of social responsiveness.

\par Finally, Praxis is assessed using the Florida Apraxia Battery (modified for children) [\cite{mostofsky2006developmental}]. It assesses the ability to perform skilled motor gestures on command, by imitation, and with actual tool use. Several studies [\cite{mostofsky2006developmental}, \cite{dziuk2007dyspraxia}, \cite{dowell2009associations}, \cite{nebel2016intrinsic}] reveal that children with ASD show marked impairments in Praxis a.k.a., developmental dyspraxia, and that impaired Praxis correlates with impairments in core autism social-communicative and behavioral features. Performance is videotaped and later scored by two trained research-reliable raters, with total percent correctly performed gestures as the dependent variable of interest. Scores therefore range from $0-100$, with higher scores indicating better Praxis performance. This measure was available for only $48$ of the $57$ subjects in the KKI dataset.

\subsection {Evaluating Predictive Performance}

\begin{figure}[b!]
   \centering
   \fbox{\includegraphics[scale=0.30]{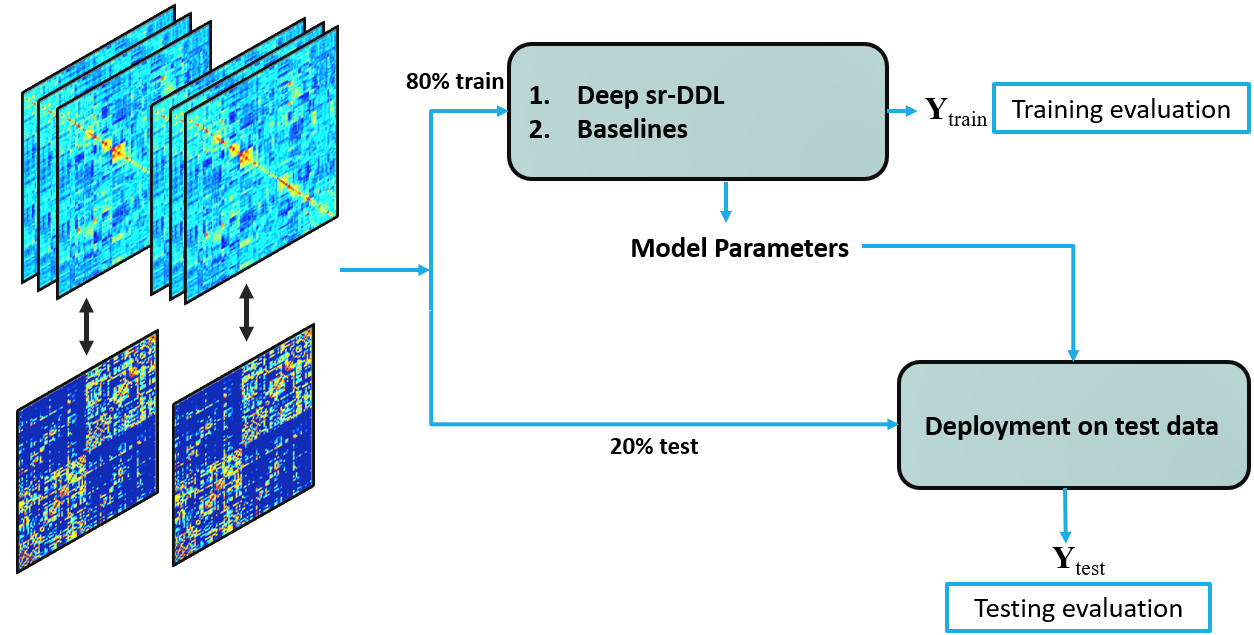}}
   \caption{A five-fold cross validation for evaluating performance}\label{fig:CV}
\end{figure}

\par {We characterize the performance of each method using a five-fold cross validation strategy, as illustrated in Fig.~\ref{fig:CV}.}

\par {We report three quantitative measures of performance. The first is the Median Absolute Error (MAE) between the outputs $\hat{\mathbf{y}}_{n}$ and the true scores $\mathbf{y}_{n}$, computed as :
\begin{equation}
\mathrm{MAE} = \mathrm{median}(\vert{\mathbf{\hat{y}}_{:,m}-\mathbf{y}_{:,m}}\vert),
\label{eqn:MAE}
\end{equation}
The MAE quantifies the absolute distance between the measured and predicted scores across individuals. We report MAE along with the corresponding standard deviation of the errors to quantify robustness.} Lower MAE indicates better testing performance.

\par The second metric is the Normalized Mutual Information (NMI), which assesses the similarity in the distribution of the predicted and observed score distributions across subjects. NMI for the score $m$ is computed as:
\begin{equation}
\mathrm{NMI}(\mathbf{y}_{:,m},\mathbf{\hat{y}_{:,m}}) = \frac{H(\mathbf{y}_{:,m}) + H(\mathbf{\hat{y}_{:,m}}) - H(\mathbf{y}_{:,m},\mathbf{\hat{y}}_{:,m})}{\min{\{H(\mathbf{y}_{:,m})},H(\mathbf{\hat{y}}_{:,m}) \}}  
\end{equation}
Here, $H(\mathbf{y}_{:,m})$ is the entropy of $\mathbf{y}_{:,m}$ and $H(\mathbf{y}_{:,m},\mathbf{\hat{y}}_{:,m})$ is the joint entropy between $\mathbf{y}_{:,m}$ and $\mathbf{\hat{y}}_{:,m}$. NMI ranges between $0-1$ with a higher value indicating better agreement between predicted and measured score distributions, and thus characterizing improved performance.

\par {Finally, we report the $R^2$ metric or the coefficient of determination evaluated on the predicted and true scores. Intuitively, the $R^2$ is a statistical measure that helps us assess the amount of variance in the true scores, i.e. $\mathbf{y}_{m}$ (for the $m^{th}$) score that is explained by the corresponding $\hat{\mathbf{y}}_{m}$ as predicted by the method. This is mathematically reported as
\begin{equation*}
    R^{2}(\mathbf{y}_{m},\hat{\mathbf{y}}_{m})= 1 - \frac{\sum_{i}{(\mathbf{y}_{m}(i) - \bar{\mathbf{y}}_{m})^2}}{\sum_{i}{(\mathbf{y}_{m}(i) - \hat{\mathbf{y}}_{m}(i))^2}}
\end{equation*}
where, $\bar{\mathbf{y}}_{m}$ indicates the mean value of the true scores $\mathbf{y}_{m}$. Larger values of $R^2$ indicate better agreement between the true and predicted scores.}

\begin{figure*}   
    \centering
      \includegraphics[width=\dimexpr \textwidth-10\fboxsep-10\fboxrule\relax]{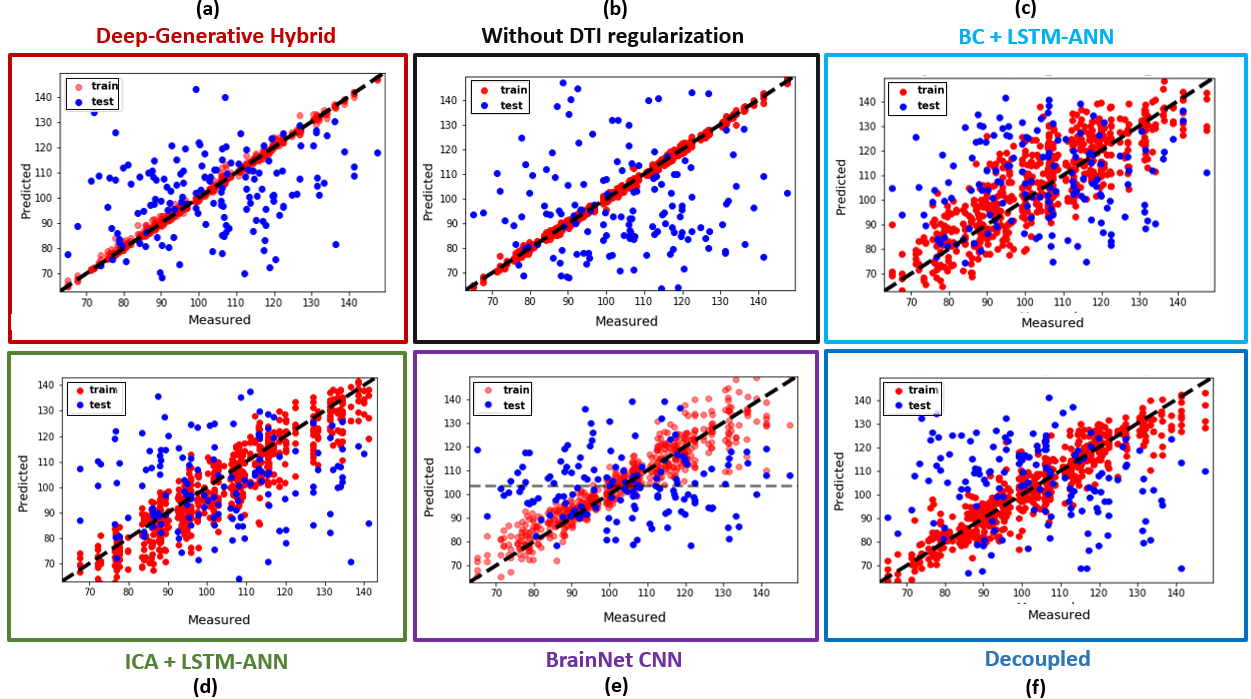}

  \caption{{\textbf{HCP dataset:} Prediction performance for the Cognitive Fluid Intelligence Score by the (a) \textbf{Red Box:} Deep sr-DDL. (b) \textbf{Black Box:} Deep sr-DDL model without DTI regularization (c) \textbf{Light Purple Box:} Betweenness Centrality on DTI + dynamic rs-fMRI multimodal graphs followed by LSTM-ANN predictor (d) \textbf{Green Box:} ICA timeseries followed by LSTM-ANN predictor (e) \textbf{Purple Box}: Branched BrainNet CNN [\cite{kawahara2017brainnetcnn}] on DTI and rs-fMRI static graphs (f) \textbf{Blue Box:} Decoupled DDL factorization followed by LSTM-ANN predictor}} 
  \label{fig:HCP}
\end{figure*}

\begin{figure*}    
    \centering
      \includegraphics[width=\dimexpr \textwidth-15\fboxsep-20\fboxrule\relax]{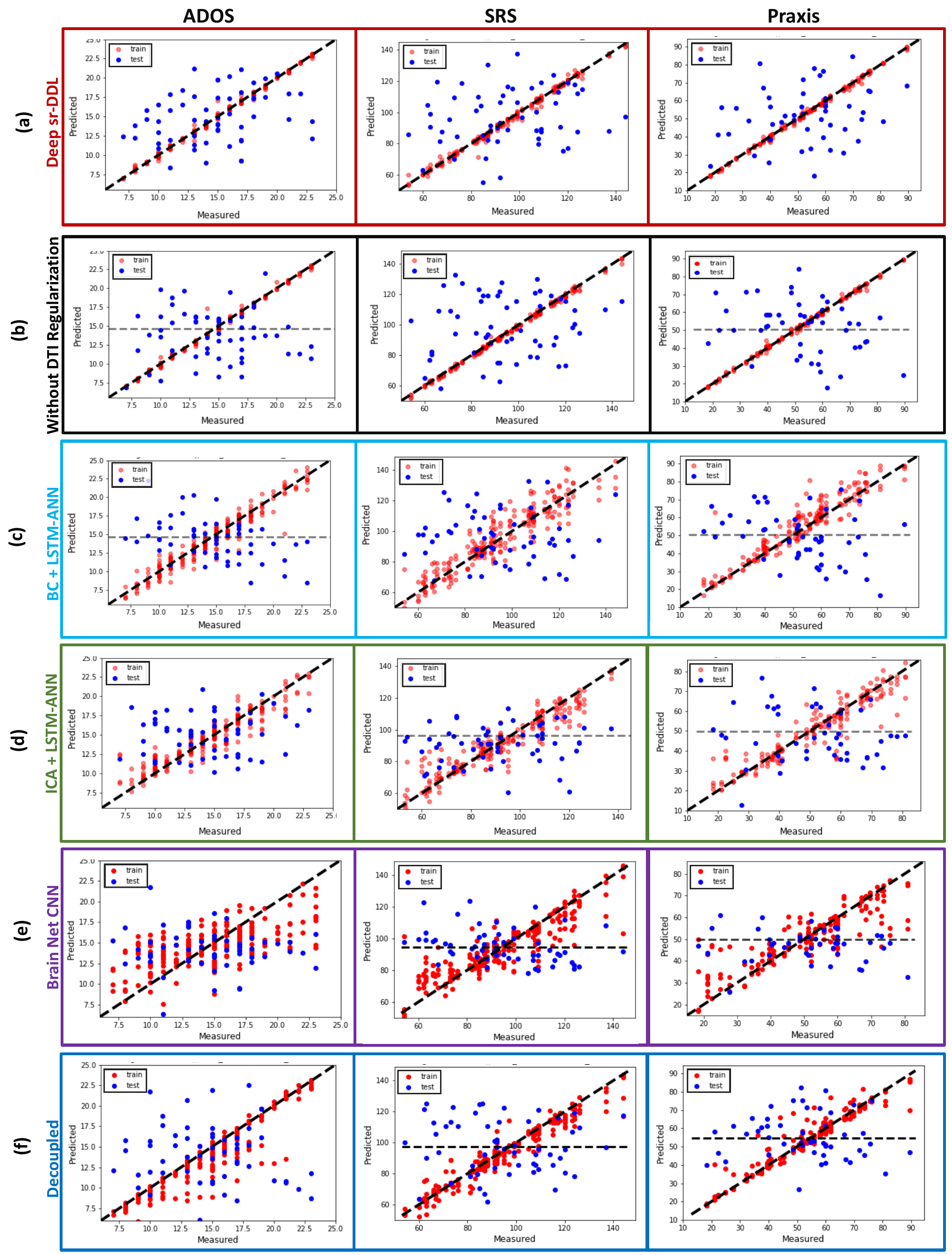}

  \caption{ {\textbf{KKI dataset:} Multiscore prediction performance for the \textbf{(L)} ADOS, \textbf{(M)} SRS, and \textbf{(R)} Praxis by the \textbf{(a) Red Box:} Deep sr-DDL \textbf{(b) Black Box:} Model without DTI regularization \textbf{(c) Light Purple Box:} Betweenness Centrality on DTI + dynamic rs-fMRI multimodal graphs followed by LSTM-ANN predictor \textbf{(d) Green Box:} ICA timeseries followed by the LSTM-ANN predictor \textbf{(e) Purple Box}: Branched BrainNet CNN [\cite{kawahara2017brainnetcnn}] on DTI Laplacian and rs-fMRI static graphs \textbf{(f) Blue Box:} Decoupled DDL factorization followed by LSTM-ANN predictor} }
  \label{fig:KKI}
\end{figure*}

\begin{table*}[h!]
\centering
{\renewcommand{\arraystretch}{1.2}
\begin{tabular}{|c |c | c| c| c| c | c |} 
\hline 
  \textbf{Score} &\textbf{Method} &\textbf{MAE Train } & \textbf{MAE Test} & \textbf{NMI Train} & \textbf{NMI Test} & $R^{2}$ \textbf{Test} \\  
\hline 
\hline
  \multirow{7}{6em}{CFIS} & Median & N/A & 13.51~\rpm~{9.97} & N/A & 0 & $1e^{-21}$\\
  & BC \& LSTM-ANN & {7.23}~\rpm~{6.24} & {16.50}~\rpm~{13.60} & 0.53 & 0.72 & 0.013 \\
 & ICA \& LSTM-ANN & {4.87}~\rpm~{4.84} & {16.45}~\rpm~{14.7} & 0.58 & \textbf{0.77} & 0.013\\
 & BrainNet CNN &  {3.50}~\rpm~{2.1} & {16.89}~\rpm~{12.20}  & 0.79 & 0.73 & 0.0017 \\
 & Decoupled & 3.72~\rpm~{4.33} & {18.10}~\rpm~{14.04} & 0.78 & 0.70 & 0.011\\
 & Without~DTI~regularization & \underline{{0.77}~\rpm~{0.66}} & {20.02}~\rpm~{15.04} & \textbf{0.88} & {0.74} & 0.0089\\
 & \textbf{Deep sr-DDL} & \textbf{{0.44}~\rpm~{0.15}}& \textbf{{14.76}~\rpm~{12.77}}& \underline{0.86}& \textbf{0.77} & \textbf{0.071} \\
 \hline
\end{tabular}
}
\caption{ { \textbf{HCP Dataset:} Performance evaluation on the HCP dataset against our prior work according to \textbf{Median Absolute Error (MAE)}, \textbf{Normalized Mutual Information (NMI)}, and $R^2$. We also report the standard deviation for the MAE Lower MAE and higher NMI/$R^2$ score indicate better performance. Best performance is highlighted in bold.}}
\label{table:HCP}
\end{table*}
\textcolor{Purple}{
\begin{table*}[b!]
\centering
{\renewcommand{\arraystretch}{1.2}
\begin{tabular}{|c |c | c| c| c| c | c|} 
\hline 
  \textbf{Score} &\textbf{Method} &\textbf{MAE Train } & \textbf{MAE Test} & \textbf{NMI Train} & \textbf{NMI Test} & \textbf{$R^2$ Test} \\  
\hline 
\hline
  \multirow{7}{4em}{ADOS} 
 & Median & N/A & 2.33~\rpm~{2.01} & N/A & 0 & $1e^{-31}$ \\
 & BC \& LSTM-ANN & {0.68}~\rpm~{0.57} & {4.36}~\rpm~{3.36} & 0.89 & 0.29 & 0.01 \\
 & ICA \& LSTM-ANN & {0.9}~\rpm~{0.54} & \textbf{{2.47}~\rpm~{2.04}} & 0.91 & \textbf{0.41} & \textbf{0.25} \\
 & BrainNet CNN & {1.90}~\rpm~{0.086} & {3.50}~\rpm~{2.20} & 0.96 & 0.25 & 0.17 \\
 & Decoupled & {1.34}~\rpm~{0.51} & {3.93}~\rpm~{2.10} & 0.68 & 0.29 & 0.06 \\
 & Without DTI regularization & 0.25~\rpm~{0.099} & {3.50}~\rpm~{3.09} & 0.99 & 0.17 & 0.02\\
 & \textbf{Deep sr-DDL} & \textbf{0.2~\rpm~{0.09}}& \underline{2.99~\rpm~{1.99}}& \textbf{0.99}& \underline{0.37} & \underline{0.23} \\
[0.2ex]  
\hline
 \multirow{7}{4em}{SRS} 
 & Median & N/A & {16.81}~\rpm~{12.8} & N/A & 0 & $1e^{-30}$ \\
 & BC \& LSTM-ANN & {5.10}~\rpm~{4.61} & \underline{{18.05}~\rpm~{14.22}} & 0.92 & \underline{0.83} & 0.09 \\
 & ICA \& LSTM-ANN & {5.27}~\rpm~{3.32} & \textbf{{13.64}~\rpm~{12.69}} & 0.76 & 0.59 & 0.008\\
 & BrainNet CNN & 5.25~\rpm~{2.5} & {18.96}~\rpm~{15.65} & 0.83 & 0.75 & 0.018 \\
 & Decoupled & {2.10}~\rpm~{2.98} & {21.45}~\rpm~{13.73} & {0.76} & 0.78 & 0.002 \\
 & Without DTI regularization & \textbf{0.72~\rpm~{0.61}} & {22.20}~\rpm~{14.78} & 0.95 & 0.65 & 0.08\\
 & \textbf{Deep sr-DDL} &  \underline{{1.21}~\rpm~{0.66}}& \underline{{18.70}~\rpm~{13.51}}& \textbf{0.98}& \textbf{0.85} & \textbf{0.12} \\ [0.2ex]
 \hline
 \multirow{7}{4em}{Praxis} 
 & Median & N/A & 10.53~\rpm~{8.81} & N/A & 0 & $1e^{-29}$ \\
 & BC \& LSTM-ANN & 6.61~\rpm~{3.30} & 17.49~\rpm~{9.08} & 0.86 & 0.70 & 0.01 \\
 & ICA \& LSTM-ANN & 4.56~\rpm~{1.26} & {15.02}~\rpm~{11.80} & 0.82 & 0.60 & 0.0122 \\
 & BrainNet CNN  & 3.78~\rpm~{0.59} & 15.15~\rpm~{11.49} & 0.95 & 0.19 & 0.009 \\
 & Decoupled & 1.57~\rpm~{1.12} & 21.67~\rpm~{12.02} & 0.75 & 0.25 & 0.003 \\
 & Without DTI regularization & \textbf{0.61~\rpm~{0.29}} & 18.56~\rpm~{14.32} & \textbf{0.96} & 0.65 & 0.08 \\ 
 & \textbf{Deep sr-DDL} &  \underline{{0.62}~\rpm~{0.36}}& \textbf{{14.99}~\rpm~{10.17}} & \underline{0.95}& \textbf{0.82} & \textbf{0.10} \\ [0.2ex]
 \hline
\end{tabular}
}
\caption{{{\textbf{KKI Dataset:} Performance evaluation on the KKI dataset against our prior work according to \textbf{Median Absolute Error (MAE)}, \textbf{Normalized Mutual Information (NMI)}, and $R^2$. We also report the standard deviation for the MAE Lower MAE and higher NMI/$R^2$ score indicate better performance. Best performance is highlighted in bold.}}}
\label{table:KKI}
\end{table*} 
}
\subsection{ Multi-Score Prediction on Real World Data}
\label{Results}
Similarly, Fig.~\ref{fig:HCP} illustrates the performance comparison of our deep sr-DDL framework against the baselines in Section~\ref{baselines} on the HCP dataset for predicting the CFIS. Fig.~\ref{fig:KKI} presents the same comparison on the KKI dataset for multi-score prediction.
In each figure, the scores predicted by the algorithm are plotted on the $\mathbf{y}$-axis against the measured ground truth score on the $\mathbf{x}$-axis. The bold $\mathbf{x}= \mathbf{y}$ line represents ideal performance. The red points represent the training data, while the Purple points indicate the held out testing data for all the cross validation folds. 

\par We observe that the training performance of the baselines is good (i.e. the red points follow the $\mathbf{x}=\mathbf{y}$ line) in all cases for both datasets. However, in case of testing performance, our method outperforms the baselines in all cases. This performance gain is particularly pronounced in the case of multiscore prediction (KKI dataset). Empirically, we are able to tune the baseline hyperparameters to obtain good testing performance on the KKI dataset for a single score (ADOS for ICA+LSTM-ANN), but the prediction of the remaining scores (SRS and Praxis for the KKI dataset) suffers. {Notice that the prediction on one or more of scores (KKI dataset) and CFIS (HCP dataset) hovers around the population median of the score in several cases. In fact, in some of the multi-score prediction cases, it performs worse than predicting the median. This is testament to the inherent difficulty of the prediction task at hand.} Finally, we notice that omitting the structural regularization from the deep sr-DDL performs worse than our method.

\par In contrast to the baselines, the testing predictions of our framework follow the $\mathbf{x}=\mathbf{y}$ more closely. The machine learning, statistical and graph theoretic techniques we selected for a comparison are well known in literature for being able to robustly provide compact characterizations for high dimensional datasets. However, we see that ICA is unable to estimate a reliable projection of the data that is particularly useful for behavioral prediction. Similarly, the betweenness centrality measure is unable to extract informative topologies for brain-behavior integration. We conjecture that the aggregate nature of this measure is useful for capturing group-level commonalities, but falls short of modeling subject-specific differences. Furthermore, even the BrainNet CNN, which directly exploits the graph structure of the connectomes falls short of generalizing to multi-score prediction. Additionally, it ignores the dynamic information in the rs-fMRI data. In case of the baseline where we omit the structural regularization, i.e. deep sr-DDL without DTI, we notice that the method learns a representation of the rs-fMRI data that generalizes beyond the training set, but still falls short of the performance when anatomical information is included. This clearly demonstrates the benefit of supplementing the functional data with structural priors. Finally, the failure of the decoupled dynamic matrix factorization and deep-network makes a strong case for jointly optimizing the neuroimaging and behavioral representations. The basis estimated independently of behavior are not indicative of clinical outcomes, due to which the regression performance suffers. We also quantify the performance indicated in these figures in Table~\ref{table:HCP} (HCP dataset) and Table~\ref{table:KKI} (KKI dataset) based on the MAE and NMI/{$R^2$. For reference, we have added an additional row as a `baseline' in our tables where for each test subject, we simply predict the median of each score.}

\par Our deep sr-DDL framework explicitly optimizes for a viable tradeoff between multimodal and dynamic connectivity structures and behavioral data representations jointly. The dynamic matrix decomposition simultaneously models the group information through the basis, and the subject-specific differences through the time-varying coefficients. The DTI Laplacians streamline this decomposition to focus on anatomically informed functional pathways. The LSTM-ANN directly models the temporal variation in the coefficients, with its weights encoding representations closely interlinked with behavior. The limited number of basis elements help provide compact representations explaining the connectivity information well. The regularization and constraints ensure that the problem is well posed, yet extracts clinically meaningful representations.
\begin{figure*}[b!]

\centering
   \includegraphics[width=\dimexpr \textwidth-10\fboxsep-10\fboxrule\relax]{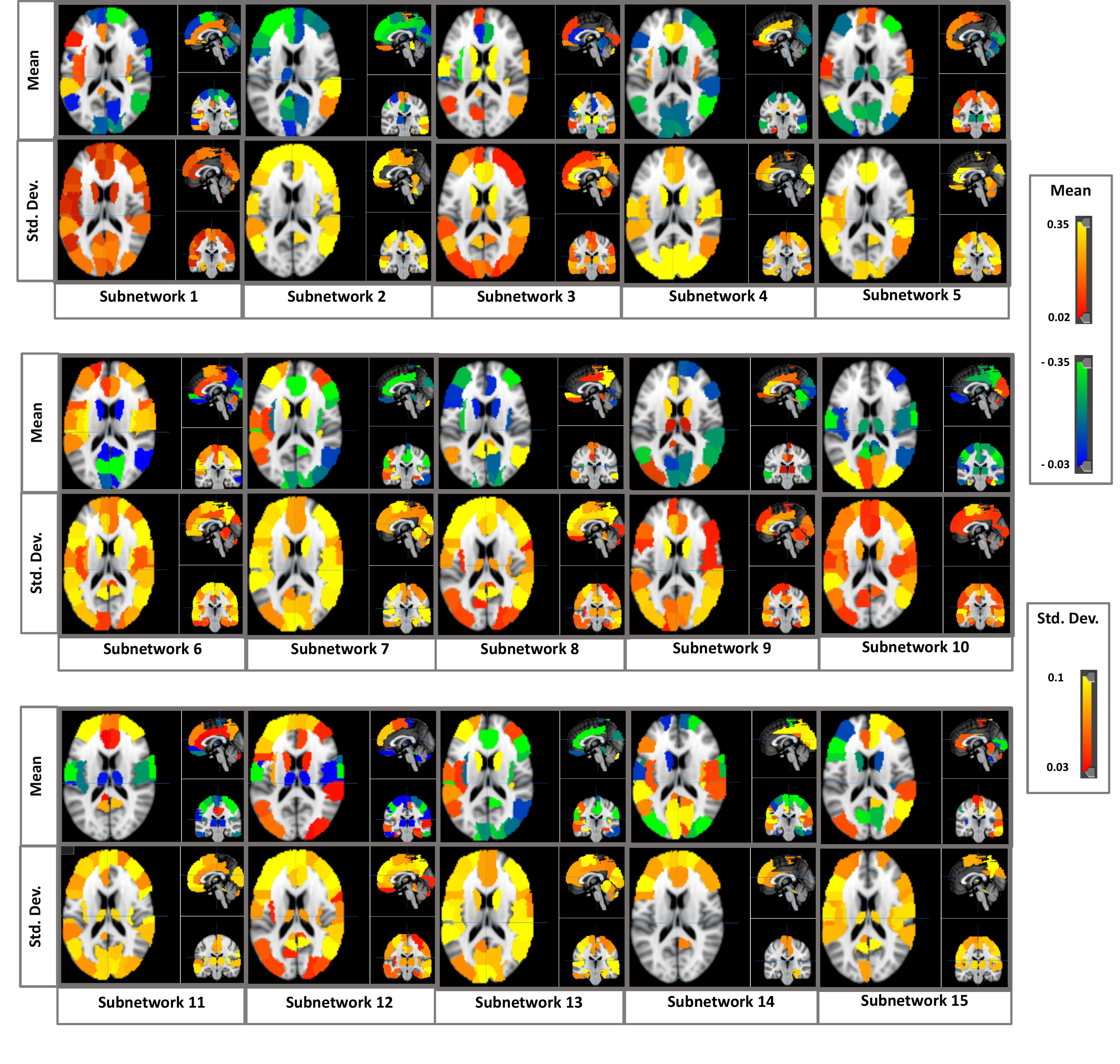}

\caption{\footnotesize{{Complete set of subnetworks identified by the deep sr-DDL model for the HCP database. \textbf{Mean}: Mean regional co-activation patterns in basis $\mathbf{B}$ The red and orange regions are anti-correlated with the Purple and green regions. \textbf{Std. Dev.}: Standard deviations of regional co-activation patterns. A majority of regions exhibit small deviations from the mean. Both sets of plots have been computed across cross-validation folds}} }
\label{fig:subnetworks_HCP} 

\end{figure*}

\begin{figure*}[h!]

\centering
   \includegraphics[width=\dimexpr \textwidth-10\fboxsep-10\fboxrule\relax]{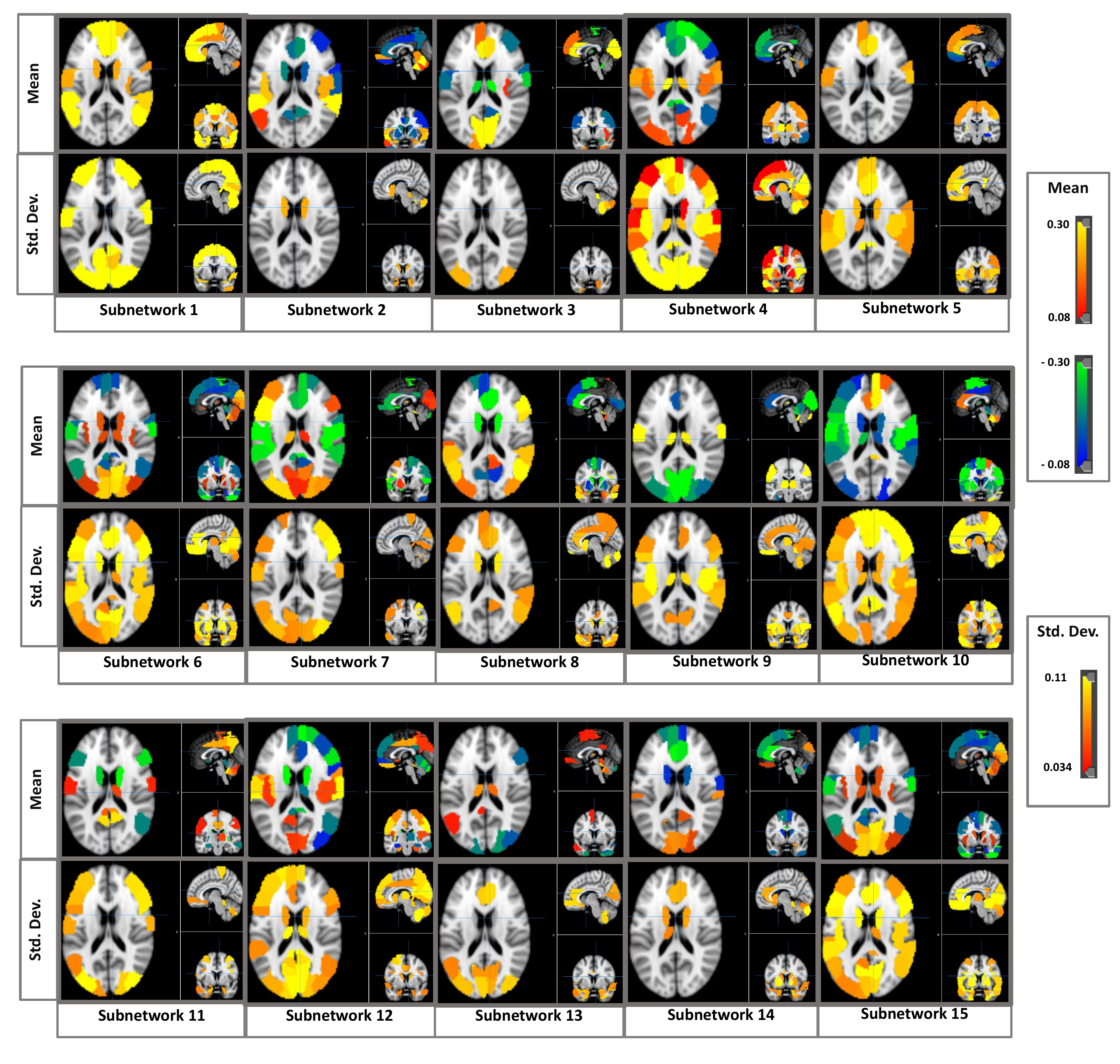}

\caption{\footnotesize{{Complete set of subnetworks identified by the deep sr-DDL model for the KKI database. \textbf{Mean}: Mean regional co-activation patterns in basis $\mathbf{B}$ The red and orange regions are anti-correlated with the Purple and green regions. \textbf{Std. Dev.}: Standard deviations of regional co-activation patterns. A majority of regions exhibit small deviations from the mean. Both sets of plots have been computed across cross-validation folds}}}
\label{fig:subnetworks} 

\end{figure*}

\subsection{Clinical Interpretation}

\paragraph{\textbf{Subnetwork Identification}} 
\label{SNI} {In this section, we investigate the subnetworks learned in the basis $\mathbf{B}$ by the sr-DDL model when trained on both datasets. Recall that each column of the basis consists of a set of co-activated AAL subregions. In order to robustly identify these patterns, we first train the model on $10$ randomly sampled subsets of each dataset.} {Then, we match the obtained subnetworks based on their absolute cosine similarity}.  {Since we have $15$ subnetworks, we then illustrate the mean co-activations across the brain regions for each of them individually in Fig.~\ref{fig:subnetworks_HCP} (HCP) and Fig.~\ref{fig:subnetworks} (KKI). Here, the colorbar in the figure indicates subnetwork contribution to the AAL regions. Regions storing negative values (cold colors) are anticorrelated with regions storing positive ones (hot colors). Alongside, we represent the corresponding standard deviations across different regions for each of the $15$ subnetworks.}

\par {Examining the subnetworks in Fig.~\ref{fig:subnetworks_HCP}, we notice that} {Subnetworks~$1$ \& $2$, and $11$} {exhibits positive and competing contributions from regions of the Default Mode Network (DMN), which has been widely inferred in the resting state literature [\cite{raichle2015brain}] and is believed to play a critical role in consolidating memory [\cite{sestieri2011episodic}], as also in self-referencing and in the theory of mind [\cite{andrews2012brain}]. At the same time,} {Subnetworks~$2$ and $11$} {have competing and positive contributions from regions in the Frontoparietal Network (FPN) respectively. The FPN is known to be involved in executive function and goal-oriented, cognitively demanding tasks [\cite{uddin2019towards}].} {Subnetworks~$1$, $6$, $7$, $11$ and $13$} {are comprised of regions from the Medial Frontal Network (MFN). The MFN and FPN are known to play a key role in decision making, attention and working memory [\cite{menon2011large,euston2012role}], which are directly associated with cognitive intelligence.} {Subnetworks~$1$, $3$, and $9$} {include contributions from the subcortical and cerebellar regions, while} {Subnetworks~$10$, $2$, $14$ and $11$} {include contributions from the Somatomotor Network (SMN).
Taken together, these networks are believed to be important functional connectivity biomarkers of cognitive intelligence and consistently appear in previous literature on the HCP dataset [\cite{hearne2016functional,chen2019resting}].}

\par {For the KKI dataset, in Fig.~\ref{fig:subnetworks}, {Subnetwork~$1$} includes regions from the DMN, and the SMN. Similarly,} {Subnetwork~$6$} {includes competing contributions from the SMN and DMN regions. Aberrant connectivity within the DMN and SMN regions have previously been reported in ASD [\cite{lynch2013default,nebel2016intrinsic}].} {Subnetworks~$7$, $3$, and $6$} {exhibit contributions from higher order visual processing areas in the occipital and temporal lobes along with competing sensorimotor regions. At the same time,} {Subnetwork~$9$} {exhibits competing contributions from the visual network. These findings concur with behavioral reports of reduced visual-motor integration in autism [\cite{nebel2016intrinsic}].} {Subnetworks~$11$ and $8$} {exhibit contributions from the central executive control network (CEN) and insula.} {Subnetwork~$10$} {also exhibits anticorrelated CEN contributions. These regions are believed to be essential for switching between goal-directed and self-referential behavior [\cite{sridharan2008critical}].} {Subnetwork~$5$ and Subnetwork~$3$} {includes prefrontal and DMN regions, along with subcortical areas such as the thalamus, amygdala and hippocampus. The hippocampus is known to play a crucial role in the consolidation of long and short term memory, along with spatial memory to aid navigation. Altered memory functioning has been shown to manifest in children diagnosed with ASD [\cite{williams2006profile}]. The thalamus is responsible for relaying sensory and motor signals to the cerebral cortex in the brain and has been implicated in autism-associated sensory dysfunction, a core feature of ASD [\cite{cascio2008tactile}]. Along with the amygdala, which is known to be associated with emotional responses, these areas may be crucial for social-emotional regulation in ASD. [\cite{pouw2013link}].}

\par {Finally, we notice that the standard deviations for a majority of the regions in each of the subnetworks are small compared to the mean coactivation. Additionally, we observed an average similarity of} {$0.79~\rpm~0.13$} {and $0.81~\rpm~0.12$ for these subnetworks across the runs on subsets of the HCP and KKI datasets respectively.} These results suggests that our deep-generative framework is able to capture stable underlying mechanisms which robustly explain the different sets of deficits in ASD as well as robustly extract signatures of cognitive flexibility in neurotypical individuals.
\paragraph{\textbf{Study of Emerging Patterns}}

\begin{figure*}[b!]

\centering
   \includegraphics[width=\dimexpr \textwidth-10\fboxsep-10\fboxrule\relax]{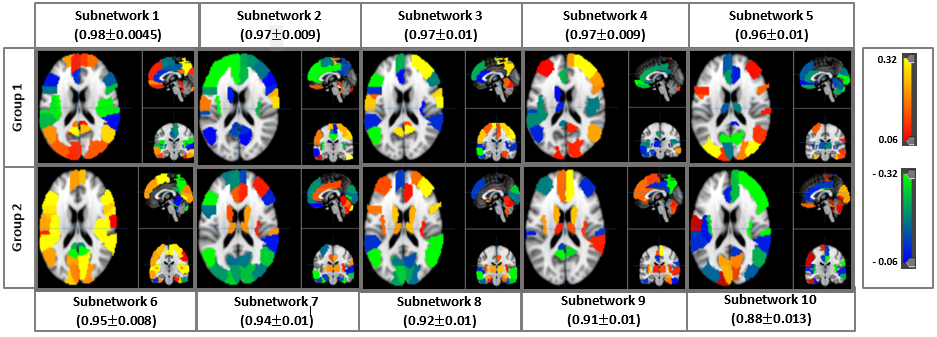}

\caption{{\footnotesize{{\textbf{HCP dataset:} Set of top $10$ consistent subnetworks across different model orders. Subnetworks in group $1$ exhibit above $0.95$ average similarity across data subsets and model orders. Subnetworks in group $2$ exhibit between $0.85-0.95$ average similarity across data subsets and model orders.}}
\label{Sup_HCP}
}}
\end{figure*}

\begin{figure*}[h!]

\centering
   \includegraphics[width=\dimexpr \textwidth-10\fboxsep-10\fboxrule\relax]{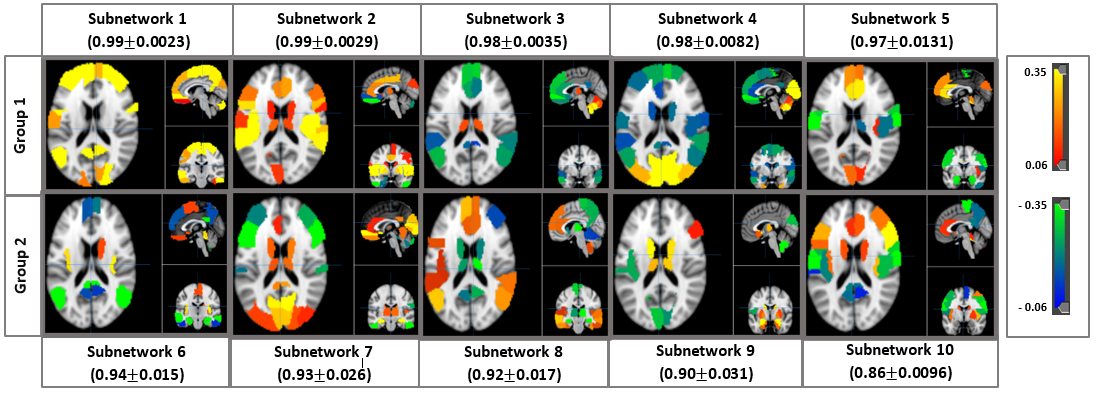}

\caption{{\footnotesize{{\textbf{KKI dataset:} Set of top $10$ consistent subnetworks across different model orders. Subnetworks in group $1$ exhibit above $0.95$ average similarity across data subsets and model orders. Subnetworks in group $2$ exhibit between $0.85-0.95$ average similarity across data subsets and model orders.}}
\label{Sup_KKI}
}}
\end{figure*}
\label{AppdD}
{
In this experiment, we study the overlap in the subnetworks in the basis $\mathbf{B}$ across different scales of subnetworks, i.e. varying the number of networks $K$. Recall from Section~\ref{Implementation}, that the knee point of the eigen-spectrum of $\{\mathbf{\Gamma}^{t}_{n}\}$ for both datasets is between $8-20$. Namely, we re-run the sr-DDL model on both the datasets steadily increasing the number of networks from $8-20$. In each case, we repeat the experiment using $10$ random subsets of the data and look for subnetworks that appear most often. Fig.~\ref{fig:subnetworks_HCP} and Fig.~\ref{fig:subnetworks} illustrate the top ten networks that appear most frequently across different data subsets and choice of $K$ for the HCP dataset and KKI dataset respectively. Alongside, we also report the mean and standard deviation of the \textcolor{Purple}{absolute cosine} similarity (S) for each individual subnetworks across the multiple runs. Networks which are most consistent exhibit higher similarity across runs with group $1$ being the top five subnetworks (S $\geq 0.95$), group $2$ being the next five subnetworks ($S>0.85$). Finally, a visual inspection and comparison with our results in Section~\ref{SNI} suggest a considerable overlap between the subnetworks in Fig.~\ref{fig:subnetworks_HCP} and Fig~\ref{Sup_HCP} for the HCP dataset and between Fig.~\ref{fig:subnetworks} and Fig~\ref{Sup_KKI} for the KKI dataset. These results suggest that our Deep sr-DDL robustly extracts representative neural signatures indicative of behavior in both healthy and autistic populations. }

\paragraph{\textbf{Decoding rs-fMRI networks dynamics}}

\begin{figure}[t!]
 \centerline{\includegraphics[scale=0.34]{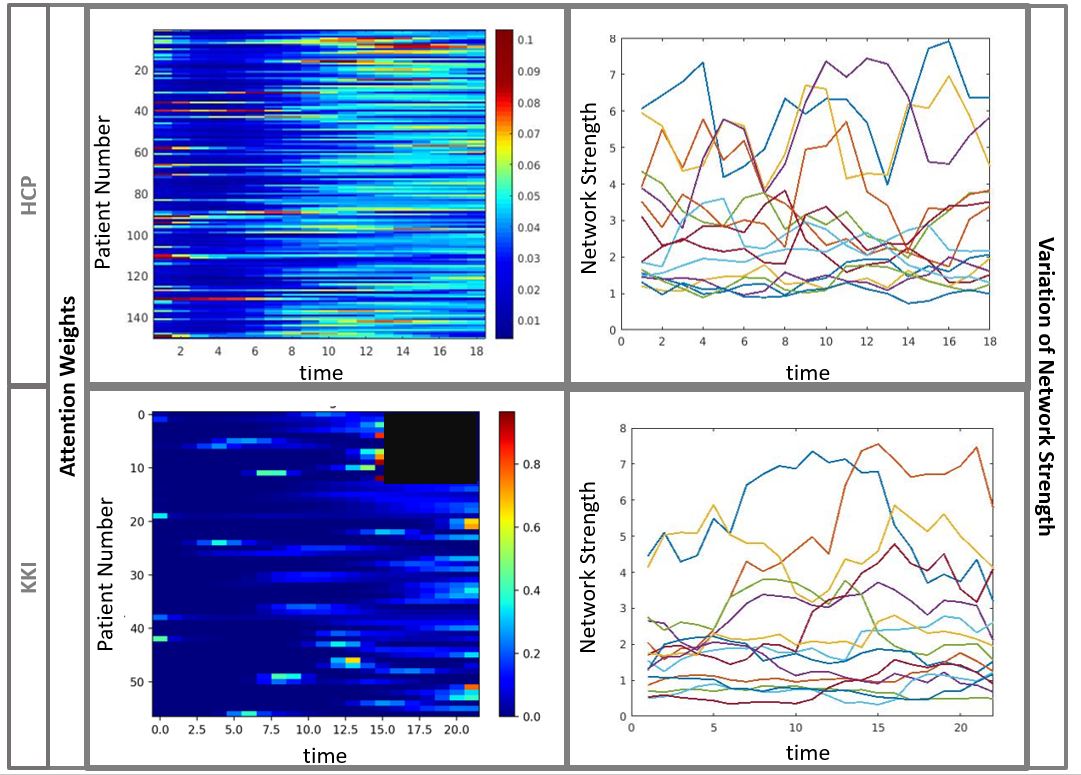}}
{\caption{ {\textbf{(Left)} Learned attention weights \textbf{(Right)} Variation of network strength over time on the \textbf{(Top)} HCP dataset \textbf{(Bottom)} KKI dataset}}
\label{fig:attn}} 
\end{figure}

Our deep sr-DDL allows us to map the evolution of functional networks in the brain by probing the LSTM-ANN representation. Recall that our model does not require the rs-fMRI scans to be of equal length. Fig.~\ref{fig:attn} (left) illustrates the learned attentions output by the A-ANN for the \textcolor{Purple}{$150$ subjects} from the HCP dataset on the top and the $57$ KKI subjects at the bottom during testing. For the KKI dataset, the patients with shorter scans have been grouped in the top of the figure. These time-points have been blackened at the beginning of the scan. The colorbar indicates the strength of the attention weights. Higher attention weights denote intervals of the scan considered especially relevant for prediction. Notice that the network highlights the start of the scan for several individuals, while it prefers focusing on the end of the scan for some others, especially pronounced in case of the KKI dataset. The patterns are comparatively more diffused for subjects in the HCP dataset, although several subjects manifest selectivity in terms of relevant attention weights. This is indicative of the underlying individual-level heterogeneity in both the cohorts.

\par Next, we illustrate the variation of the network strength for a representative subject from the HCP dataset and KKI dataset over the scan duration in Fig.~\ref{fig:attn} (right) at the top and bottom respectively. Each solid colored line corresponds to one of the $15$ sub-networks in Fig.~\ref{fig:subnetworks}. Notice that, over the scan duration, each network cycles through phases of activity and relative inactivity. Consequently, only a few networks at each time step contribute to the patient's dynamic connectivity profile. This parallels the transient brain-states hypothesis in dynamic rs-fMRI connectivity [\cite{allen2014tracking}], with active states as corresponding sub-networks in the basis matrix $\mathbf{B}$.

\section{Discussion}

Our deep-generative hybrid cleverly exploits the intrinsic structure of the rs-fMRI correlation matrices through the dynamic dictionary representation to simultaneously capture group-level and subject-specific information. At the same time, the LSTM-ANN network models the temporal evolution of the rs-fMRI data to predict behavior. The compactness of our representation serves as a dimensionality reduction step that is related to the clinical score of interest, unlike the pipelined treatment commonly found in the literature. Our structural regularization helps us fold in anatomical information to guide the functional decomposition. Overall, our framework outperforms a variety of state-of-the-art graph theoretic, statistical and deep learning baselines on two separate real world datasets.

\par  We conjecture that the baseline techniques fail to extract representative patterns from structural and functional data. These techniques are quite successful at modelling group level information, but fail to generalize to the entire spectrum of cognitive, symptomatic or connectivity level differences among subjects. Consequently, they overfit the training data.  
\begin{figure}[h!]    
    \centering
   \includegraphics[scale=0.42]{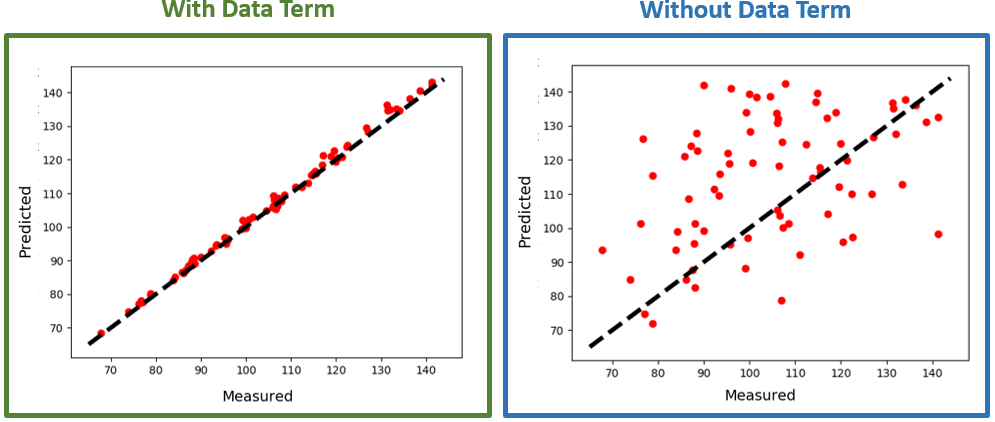}
   \caption{{Prediction Performance of the Deep sr-DDL for the CFIS score on training data when \textbf{(L)} The data term is included in computing $\{\mathbf{c}^{t}_{n}\}$ \textbf{(R)} The data term is excluded from the computation of $\{\mathbf{c}^{t}_{n}\}$ }}
   \label{OverP}
\end{figure}
\begin{figure}[b!]
 \centerline{
 
 \includegraphics[scale=0.40]{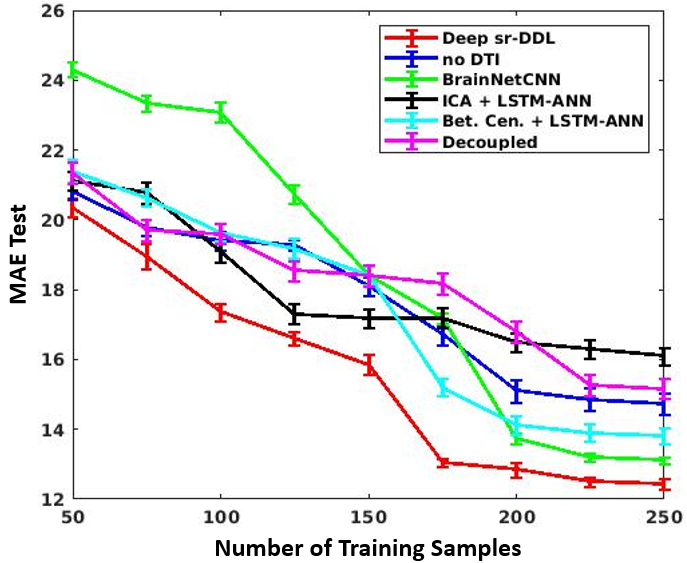}}
{\caption{{Median Absolute Error on the Test Set varying the number of samples used for training. The vertical bars indicate standard errors for each setting}}
\label{fig:test}} 
\end{figure}
\subsection{Examining Generalizability}  
\label{Gen}
\par {Notice that the training examples (red points) in Figs.~\ref{fig:HCP} and \ref{fig:KKI} follow the $\mathbf{x}=\mathbf{y}$ line perfectly, which may suggest overfitting. This phenomenon can be explained by the difference between our training procedure, where we optimize our joint objective in Eq.~(\ref{eqn:JointObj}) assuming the scores are known, and our testing procedure. Recall that Section~$2.2$ describes the procedure for calculating the temporal sr-DDL loadings for an unseen patient i.e. $\bar{\mathbf{c}}^{t}_{n}$ from the basis $\mathbf{B}^{*}$ obtained during training. Since the subject is not a part of the training set, the corresponding value of $\mathbf{\hat{y}}$ is unknown. Effectively, we must set the contribution from the data term, i.e., the deep network loss $\mathcal{L}(\cdot)$ in Eq.~(\ref{eqn:JointObj}) to $0$. Here, we examine the effect of employing the same strategy to calculate the coefficients for the training patients. In essence, we estimate the corresponding severity $\hat{\mathbf{Y}}$ now excluding the deep network loss. Accordingly, Fig.~\ref{OverP} highlights the differences in training fit with and without this term included in estimating $\{\mathbf{c}^{t}_{n}\}$ for the HCP dataset. Notice that in the latter, the training accuracy for the CFIS score has the same distribution as the testing points in Fig.~\ref{fig:HCP}. In contrast, inclusion of the deep network loss in our coupled optimization overparamterizes the search space of solutions for $\{\mathbf{c}^{t}_{n}\}$ to yield a near perfect fit. }

\par {To further probe the generalization capabilities of our Deep sr-DDL, we examine the effect of training the models on different sized datasets. For this experiment, we first set aside $50$ individuals from the HCP database as a test set on which we evaluate the generalization performance. We then sweep the training set size from $N=50-200$ in increments of $25$ subjects. To avoid biasing the results, none of these subjects overlap with the HCP-$2$ validation set used for parameter tuning in Section~\ref{Implementation}. For each training set size, we randomly sample the subjects $10$ times and compute the generalization performance on the held-out set.}

{Fig.~\ref{fig:test} displays the MAE of the CFIS score prediction on the test set as a function of the training set size. As expected, we observe that with increasing training data, the performance on the test set improves at first but eventually saturates for all methods. This is evinced by a lowering of the MAE in the initial parts of the curve followed by a subsequent plateau at roughly $150-200$ samples. Based on these results, we conjecture that further addition of training data does not substantially improve the generalization capabilities of our model or the baselines. We also note that the deep sr-DDL outperforms the baselines across the entire regime. In conjunction with our results from Section~\ref{Expt}, we conclude that the deep sr-DDL model performs reasonably well for small to moderately sized datasets. This is especially important against the backdrop of potential clinical applications, many of which have datasets of modest sizes.}

\subsection{Assessing Model Robustness}
\label{Sensitivity}

\begin{figure*}[t!]
   \centering
   \includegraphics[width=\textwidth- 2pt] {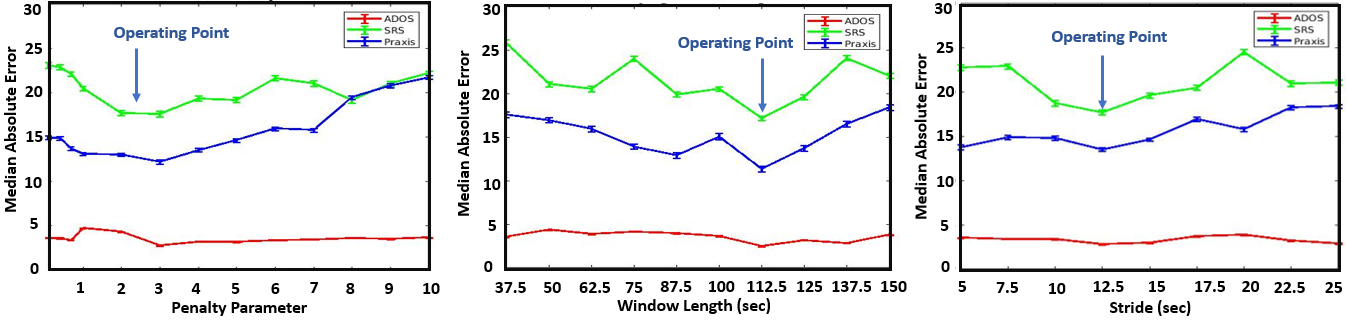}
   \caption{Performance of the Deep sr-DDL upon varying \textbf{(L):} the penalty parameter $\lambda$ \textbf{(B):} window length \textbf{(R):} stride. Our operating point is indicated by the Purple arrow}
   
   \label{fig:MitPar}

\end{figure*}

Our deep sr-DDL framework has only two free hyperparameters. The first is the number of subnetworks in $\mathbf{B}$. As described in Section~\ref{Implementation}, we use the eigen-spectrum of $\{\mathbf{\Gamma}^{t}_{n}\}$ to fix this at $15$ for both datasets. The second is the penalty parameter $\lambda$, which controls the trade-off between representation and prediction. {Recall that our data pre-processing includes a sliding window protocol in Fig.~\ref{Ex_Dyn}, which is defined by two parameters, i.e. the sliding window length and the stride. From a mathematical perspective, our deep sr-DDL formulation as such is agnostic to these parameters, as they are simply folded into the input data dimension. However, empirically, they balance the context size and information overlap within the rs-fMRI correlation matrices $\{\mathbf{\Gamma}^{t}_{n}\}$ and affects the prediction performance.}

\par In this section, we evaluate the performance of our framework under three scenarios. Specifically, we sweep $\lambda$, the window length and the stride parameter independently, keeping the other two values fixed. We use five fold cross validation with the MAE metric to quantify the multi-score prediction performance, which as shown in Section~\ref{Expt}, is more challenging than single score prediction. Fig.~\ref{fig:MitPar} plots the performance for the three scores on the KKI dataset with MAE value for each score on the $\mathbf{y}$ axis and the parameter value on the $\mathbf{x}$ axis.

\par We observed that our method gives stable performance for fairly large ranges of each parameter settings. As expected, low values of $\lambda$ $(0.01-1)$ result in higher MAE values, likely due to underfitting. Similarly, higher values $(>6)$ result in overfitting to the training dataset, degrading the generalization performance. Additionally, lower values of window lengths result in higher variance among the correlation values due to noise, and hence less reliable estimates of dynamic connectivity [\cite{lindquist2016dynamic}]. On the other hand, very large context windows tend to miss nuances in the dynamic evolution of the scan. Empirically, we observe that a mid-range of window length $100-125$s yields a good tradeoff between representation and prediction. The training of LSTM networks with very long sequence lengths is known to be particularly challenging owing to vanishing/exploding gradient issues during backpropagation. However, having too short a sequence confounds a reliable estimation of the LSTM weights from limited data. The stride parameter helps mitigate these issue by compactly summarizing the information in the sequence while simultaneously controlling the overlap across subsequent samples. Our experiments found a stride length between $10-20$s to be suitable for our application.

\par In summary, the guidelines we identified for each of the parameters are- $\lambda \in (2-5)$, window length $\in (100-125)$s, and stride $\in (10-20)$s. Additionally, our experiments on the HCP dataset using the same settings indicate that the results of our method are reproducible across different populations. {It is also interesting to note that previous experiments on the HCP dataset in literature have found similar window lengths to be stable in classification [\cite{gadgil2020spatio}] and various test-retest settings [\cite{savva2019assessment}].}

\subsection{Clinical Relevance}

{Our experiments on the KKI dataset evaluate the ability of our Deep sr-DDL framework to simultaneously explain multiple clinical impairments of ASD. This multi-target prediction is a challenging task, and in fact, the baseline methods fail to generalize all three scores. At the same time, one could evaluate the performance of predicting each score independently via three single-target regression tasks. Accordingly, Table~\ref{table:KKIssms} compares the performance of our Deep sr-DDL framework in the single-target and multi-target settings. Empirically, we observe that the single-target prediction is slightly better than the multi-target prediction. Indeed, a possible counter perspective would be to optimize for prediction accuracy of individual measures explained by potentially different brain bases, for example, as in the work of [\cite{d2019coupled}]. This comparison poses a more philosophical question about the benefits of a multi-target setup given a possible decline in predictive performance and the difficultly of the task itself.}
\begin{table}[t!]
\centering
\small
{
\begin{tabular}{|c|c|c|c|c|} 
\hline 
  \textbf{Score}&\textbf{Method} & \textbf{MAE}&\textbf{NMI} & \textbf{$R^2$}\\  
\hline 
\hline
  \multirow{2}{3em}{ADOS} 
 & Single-target & 2.91~\rpm~{2.71}& 0.44 & 0.041 \\
  & Multi-target & 2.99~\rpm~{1.99}& 0.37 & 0.23 \\
[0.2ex]  
\hline
 \multirow{2}{3em}{SRS} 
 & Single-target &  {14.78}~\rpm~{14.24} & 0.87 & 0.13 \\ & Multi-target & {18.70}~\rpm~{13.51}& 0.85 & 0.12 
 \\
 [0.2ex]
 \hline
 \multirow{2}{3em}{Praxis} 
 & Single-target &  {12.40}~\rpm~{11.60} &  0.85 & 0.06 \\
 & Multi-target &{14.99}~\rpm~{10.17} &  0.82 & 0.10\\
 [0.2ex]
 \hline
\end{tabular}
}
\caption{{Testing performance (5-fold CV) of the sr-DDL framework for single-target and multi-target prediction on the KKI dataset according to \textbf{Median Absolute Error (MAE)}, \textbf{Normalized Mutual Information (NMI)}, and $R^2$. We also report the standard deviation for the MAE. Lower MAE and higher NMI/$R^2$ scores indicate better performance.} }
\label{table:KKIssms}
\end{table} 

\par {To weigh in on this trade off, we note the growing consensus in clinical psychiatry that complex disorders, such as autism and schizophrenia, are inherently multidimensional [\cite{havdahl2016multidimensional}]. Furthermore, there is considerable patient heterogeneity within a single diagnostic umbrella that reflect subtle differences in the underlying etiology [\cite{hong2018multidimensional}]. In fact, the National Institute of Mental Health (NIHM) in the United States has released the RDoc research framework [\cite{insel2014nimh}], which advocates for a multidimensional characterization to understand the full spectrum of mental health and illness. In this context, our Deep sr-DDL approach provides a flexible tool to map multiple measures via a consistent and stable brain basis (as shown by the results in Section~\ref{SNI}). Thus, we view it as an important foundation to parse complex spectrum disorders that may even spur new analytical directions in brain connectomics.}

\par{Finally, our Deep sr-DDL framework is carefully designed to extract subject-level dynamic information. Namely, the attention mechanism automatically highlights portions of the rs-fMRI scan that are important for clinical prediction (Fig~\ref{fig:attn}). In fact, a comparison of the attention weights in Fig.~\ref{fig:attn} suggests considerable inter-patient variability of the intervals used for multi-target prediction in the KKI dataset, as opposed to the relatively consistent attention weights in the HCP dataset. This pattern may be linked to the heterogeneity of ASD described above. In conjunction, we observe the subnetwork contributions phasing in and out prominence over the course of the scan, which is consistent with the transient brain state hypothesis [\cite{allen2014tracking}]}

{In summary, the blend of classical generative modeling and deep learning prediction in our Deep sr-DDL framework allows for a finer-grained characterization of connectivity and behavior. Overall, we believe that the robustness, stability, clinical interpretability, and flexibility of our Deep sr-DDL render it a novel and valuable tool for the research community.
}

\subsection{Applications, Limitations and Future Scope}

As seen in our experiments in Section~\ref{Results}, our method is able to extract key predictive resting state biomarkers from healthy and autistic populations. Additionally, our deep sr-DDL makes minimal assumptions. Provided we have access to a set of consistently defined structural and functional connectivity measures and clinical scores, this analysis can be easily adapted to other neurological disorders and even predictive network models outside the medical realm. Overall, these findings broaden the scope of our method for future applications.

\par {Although we outperform several baselines on two separate datasets, our prediction performance in Section~\ref{Results} is far from perfect. This underscores that multi-score prediction is a challenging clinical problem. One of the key reasons can be attributed to inherent noise in the clinical measures themselves. For example, SRS is based on a parent-teacher questionnaire, which tends to be more subjective than a clinical exam. This renders the behavioral prediction task especially challenging, which partially accounts for the poor performance of several baselines we compared against. Keeping this in mind, a natural clinical direction of exploration is to adopt our method to predicting measures more directly related to functional connectivity, as opposed to those relying on clinical reports. Another avenue of exploration includes examining more coarse indicators of behavior, such as ordered levels of impairment instead of continuous measures (an ordinal regression problem), or the prevalence of ASD sub-types.}

\par {Another limitation to our method lies in the fact that our estimate of dynamic functional connectivity relies on the availability of a reliable sliding-window protocol. As illustrated in Section~\ref{Sensitivity}, an inappropriate window-length and stride choice has a direct bearing on the predictive performance. Moreover, this tradeoff is difficult to quantify and correct for analytically. Keeping this in mind, we are motivated to explore alternatives to the sliding window for better estimating dynamic functional connectivity, which can at the same time be robustly integrated into  multimodal data-analysis frameworks such as ours.}

\par {From the methodological standpoint, we recognize that our model is simplistic in its assumptions, particularly in the sr-DDL formulation. The DTI priors guide a data-driven classical rs-fMRI matrix decomposition in a regularization framework. This modeling choice was deliberately employed to preserve interpretability in the basis and simplify the inference procedure. A key limitation of this approach is that it does not directly consider multi-stage pathways, which may be an important mediator of functional relationships between communicating sub-regions. To this end, graph neural networks have shown great promise in brain connectivity research due to their ability to capture subtle and multi-stage interactions between communicating brain regions while exploiting the underlying hierarchy of brain organization. Consequently, these methods are emerging as important tools to probe complex pathologies in brain functioning and diagnose neurodevelopmental disorders [\cite{anirudh2019bootstrapping,parisot2018disease}]. In the future, we are exploring end-to-end graph convolutional networks that model the evolution of rs-fMRI signals on the anatomical DTI graphs.}

\section {Conclusion}
We have introduced a novel deep-generative framework to integrate complementary information from the functional and structural neuroimaging domains, which simultaneously maps to behavior. Our unique structural regularization elegantly injects anatomical information into the rs-fMRI functional decomposition, thus providing us with an interpretable brain basis. Our deep network (LSTM-ANN) not only models the temporal variation among individuals, but also helps isolate key dynamic resting-state signatures, indicative of clinical/cognitive impairments. {Our coupled optimization procedure} ensures that we learn effectively from limited training data while generalizing well to unseen subjects. Finally, our framework makes very few assumptions and can potentially be applied to study other neuropsychiatric disorders (eg. ADHD, Schizophrenia) as an effective diagnostic tool.

\medskip
\paragraph{\textbf{Acknowledgements}}
This  work  has generously been supported by the National Science Foundation CRCNS award 1822575 and CAREER award 1845430, the National  Institute  of Mental Health (R01 MH085328-09, R01 MH078160-07, K01 MH109766 and R01 MH106564), the National Institute of Neurological Disorders and Stroke (R01NS048527-08), and the Autism Speaks foundation.

\section*{Appendix A}
\label{AppdA}

{Here, we provide the detailed derivations for the Weighted Frobenius Norm expression in Eq.~(\ref{eqn:err2}). We begin with the formulation in Eq.~(\ref{eqn:WFN}) below:
\begin{equation}
    \vert\vert{\mathbf{\Gamma}^{t}_{n} - {\mathbf{B}\textbf{diag}({\mathbf{c}^{t}_{n}})\mathbf{B}^{T}}}\vert\vert_{\mathbf{L}_{n}} = 
    \vert\vert{\mathbf{E}^{t}_{n}}\vert\vert_{\mathbf{L}_{n}} 
\end{equation}
Here, $\mathbf{E}^{t}_{n}$ represents the reconstruction error in the correlation matrix $\mathbf{\Gamma}^{t}_{n}$ for patient $n$ at time $t$. For the DTI graph $\mathcal{G} =(\mathcal{V},\mathcal{E})$ for patient $n$,   $\mathbf{L}_{n} = \mathbf{V}_{n}^{-\frac{1}{2}}(\mathbf{V}_{n}-\mathbf{A}_{n})\mathbf{V}_{n}^{-\frac{1}{2}}$ is the DTI Graph Laplacian, where $\mathbf{V}_{n} = \mathbf{diag}(\mathbf{A}_{n}\mathbf{1})$ is the degree matrix and $\mathbf{1}$ is the vector of all ones. For notational convenience, we will drop the subscripts $n$ and $t$ from the following computation.}

{\begin{align*}
    \vert\vert{\mathbf{E}}\vert\vert_{\mathbf{L}} &= \Tr[{\mathbf{E}^{T}\mathbf{L}\mathbf{E}}]
    = \Tr[{\mathbf{E}^{T}{\mathbf{V}^{-\frac{1}{2}}}(\mathbf{V}-\mathbf{A}){\mathbf{V}^{-\frac{1}{2}}}\mathbf{E}}] \\[1ex]
    &= \Tr[{\tilde{\mathbf{E}}^{T}(\mathbf{V}-\mathbf{A})\tilde{\mathbf{E}}}]  \ \ \text{where} \ \  \tilde{\mathbf{E}} = \mathbf{V}^{-\frac{1}{2}}\mathbf{E} \\[1ex]
    &= \sum_{i}\sum_{j}\sum_{k} \tilde{\mathbf{E}}(i,j)[\mathbf{V}(i,k)-\mathbf{A}(i,k)] \tilde{\mathbf{E}}(k,j) \\[1ex]
    &= \sum_{i,j,k} \mathbf{V}(i,k)\tilde{\mathbf{E}}(i,j)\tilde{\mathbf{E}}(k,j)  -  \sum_{i,j,k} \mathbf{A}(i,k)\tilde{\mathbf{E}}(i,j)\tilde{\mathbf{E}}(k,j) \\[1ex]
    &= \sum_{i,j} \mathbf{V}(i,i)\tilde{\mathbf{E}}(i,j)\tilde{\mathbf{E}}(i,j)  -  \sum_{i,j,k} \mathbf{A}(i,k)\tilde{\mathbf{E}}(i,j)\tilde{\mathbf{E}}(k,j)\\[1ex]
    &= \sum_{j}\sum_{(i,k) \in \mathcal{E}} 2[\tilde{\mathbf{E}}(i,k)]^{2}  -  \sum_{j}\sum_{(i,k) \in \mathcal{E}} 2[\tilde{\mathbf{E}}(i,j)\tilde{\mathbf{E}}(k,j)] \\[1ex]
    &= \sum_{j}\Big[\sum_{(i,k) \in \mathcal{E}} [\tilde{\mathbf{E}}(i,k)]^{2} + \sum_{(i,k) \in \mathcal{E}} [\tilde{\mathbf{E}}(k,j)]^{2} \Big] \end{align*}
    \begin{align*}
    & -\sum_{j}\sum_{(i,k) \in \mathcal{E}} 2[\tilde{\mathbf{E}}(i,j)\tilde{\mathbf{E}}(k,j)] \\
    &= \sum_{j}\sum_{(i,k) \in \mathcal{E}}\Big[\tilde{\mathbf{E}}(i,j)- \tilde{\mathbf{E}}(k,j)\Big]^2 \\
    &= \sum_{(i,k) \in \mathcal{E}}\vert\vert\tilde{\mathbf{E}}(i,:)- \tilde{\mathbf{E}}(k,:)\vert\vert^2_{2} \\[1ex]
    &= {\sum_{(i,k) \in \mathcal{E}}\vert\vert{[\mathbf{V}(i,i)]^{-\frac{1}{2}}\mathbf{E}(i,:)- [\mathbf{V}(k,k)]^{-\frac{1}{2}}\mathbf{E}(k,:)}\vert\vert^{2}_{2}}
\end{align*}}

{Writing out the appropriate subscripts and superscripts we dropped earlier, we obtain the expression in Eq.~(\ref{eqn:err2}):
\begin{align*}
    \vert\vert{\mathbf{\Gamma}^{t}_{n} - {\mathbf{B}\textbf{diag}({\mathbf{c}^{t}_{n}})\mathbf{B}^{T}}}\vert\vert_{\mathbf{L}_{n}} 
    &= \sum_{(i,k)\in \mathcal{E}}{\vert\vert\tilde{\mathbf{E}}^{t}_{n}(i,:) 
    -\tilde{\mathbf{E}}^{t}_{n}(k,:)\vert\vert}_{2}^2 \\[1ex]
    &= \sum_{(i,k) \in \mathcal{E}}\vert\vert{[\mathbf{V}_{n}(i,i)]^{-\frac{1}{2}}\mathbf{E}^{t}_{n}(i,:)} \\[1ex]
    & - {[\mathbf{V}_{n}(k,k)]^{-\frac{1}{2}}\mathbf{E}(k,:)}\vert\vert^2_{2}
\end{align*}}

\section*{Appendix B}
\label{AppdB}
{In this section, we detail the calculations from Section~\ref{Optim}. Thus, our alternating minimization steps are explained as:
\paragraph{\textbf{{Step~{1}: Closed form solution for $\mathbf{B}$}}} Notice that Eq.~(\ref{eqn:const}) reduces to the following quadratic form in $\mathbf{B}$:
\begin{equation}
\mathbf{B}^{*} = \argmin_{\mathbf{B} : \ \mathbf{B}^{T}\mathbf{B}=\mathcal{I}_{K}}{{\vert\vert{\mathbf{M}-\mathbf{B}}\vert\vert}^{2}_{F}}
\end{equation}
where $\mathbf{M}$ is computed as:
\begin{multline}
    \mathbf{M} = \sum_{n}{\frac{1}{T_{n}}}\sum_{t}{(\mathbf{\Gamma}^{t}_{n}\mathbf{L}_{n}+ \mathbf{L}_{n}\mathbf{\Gamma}^{t}_{n})\mathbf{D}^{t}_{n}} +  \\ \sum_{n}{\frac{1}{T_{n}}}\Big[\sum_{t}{\frac{\gamma}{2}\mathbf{D}^{t}_{n}\mathbf{diag}(\mathbf{c}^{t}_{n})+ \gamma\mathbf{\Lambda}^{t}_{n}\mathbf{diag}(\mathbf{c}^{t}_{n})}\Big] 
\end{multline} 
We know that $\mathbf{B}$ has a closed-form Procrustes solution [\cite{everson1998orthogonal}] computed as follows. Given the singular value decomposition $\mathbf{M} = \mathbf{U}\mathbf{S}\mathbf{V}^{T}$, we have:
\begin{equation*}
    \mathbf{B}^{*} = \mathbf{U}\mathbf{V}^{T}
\end{equation*}
In essence, $\mathbf{B}$ spans the anatomically weighted space of subject-specific dynamic correlation matrices. }

{\paragraph{\textbf{Step~{2}: Updating the sr-DDL loadings  $\{\mathbf{c}_{n}^{t}\}$}} The objective $\mathcal{J}_{c}$ in Eq.~(\ref{eqn:const}) decouples across subjects. We can also incorporate the non-negativity constraint $\mathbf{c}^{t}_{nk} \geq 0$ by passing an intermediate vector $\hat{\mathbf{c}}^{t}_{n}$ through a ReLU. Thus:
\begin{equation}
     \mathbf{c}_{n}^{t} = ReLU({\mathbf{\hat{c}}_{n}}^{t})
\end{equation}
The ReLU pre-filtering allows us to optimize an unconstrained version of Eq.~(\ref{eqn:const}), as follows:
\begin{multline}
\mathcal{J}_{\hat{c}} =  \lambda \sum_{n}\mathcal{L}(\mathbf{\Theta},\{{\mathbf{c}}^{t}_{n}\};\mathbf{y}_{n}) \\ 
+ \sum_{n,t}{\frac{\gamma}{T_{n}}\Big[{\Tr{\left[{(\mathbf{\Lambda}^{t}_{n})^{T}({\mathbf{D}^{t}_{n}-\mathbf{B}\mathbf{diag}({\mathbf{c}}^{t}_{n})})}\right]}}\Big]} \\ + \sum_{n.t}{\frac{\gamma}{T_{n}}}{\Big[{{\frac{1}{2}}~{\vert\vert{\mathbf{D}^{t}_{n}-\mathbf{B}\mathbf{diag}({\mathbf{c}}^{t}_{n})}\vert\vert}_{F}^{2}}}\Big] \\
\label{eqn:unconst}
\end{multline}
This optimization can be performed via the stochastic ADAM algorithm [\cite{kingma2015adam}] by backpropagating the gradients from the loss in Eq.~(\ref{eqn:unconst}) upto the input $\{\hat{\mathbf{c}}^{t}\}$. Experimentally, we set the initial learning rate to be $0.02$, scaled by $0.9$ per $10$ iterations. Essentially, this optimization couples the parametric gradient from the Augmented Lagrangian formulation with the backpropagated gradient from the deep network (parametrized by fixed $\mathbf{\Theta}$). After convergence, the thresholded loadings ${\mathbf{c}}^{t}_{n} = ReLU(\hat{\mathbf{c}}^{t}_{n})$ are used in the subsequent steps of the minimization. }

{\paragraph{\textbf{Step~{3}: Updating the Deep Network weights-$\mathbf{\Theta}$}}  We use backpropagation on the loss $\mathcal{L(\cdot)}$ to solve for the unknowns $\mathbf{\Theta}$. Notice that we can handle missing clinical data by dropping the contributions of the unknown value of $\mathbf{y}_{nm}$ to the network loss during backpropagation. Again, we use the ADAM optimizer [\cite{kingma2015adam}] with random initialization at the first main iteration of alternating minimization. We employ a learning rate of $0.2e^{-4}$, scaled by $0.95$ every $5$ epochs, and batch-size $1$. Additionally, we train the network only for $60$ epochs to avoid overfitting.}

{\paragraph{\textbf{Step~{4}: Updating the Constraint Variables $\{\mathbf{D}^{t}_{n},\mathbf{\Lambda}^{t}_{n}\}$}}
Each of the primal variables $\{\mathbf{D}^{t}_{n}\}$ has a closed form solution given by: 
\begin{equation}
[\mathbf{D}^{t}_{n}]^{k} = \mathbf{K}\mathbf{F} 
\label{eqn:primal}
\end{equation}
where, $\mathbf{K} =  (\mathbf{diag}(\mathbf{c}_{n})\mathbf{B}^{T}+ \mathbf{\Gamma}^{t}_{n}\mathbf{L}_{n}\mathbf{B} +\mathbf{L}_{n}\mathbf{\Gamma}^{t}_{n}\mathbf{B} - \gamma \mathbf{\Lambda}_{n})$ and $
\mathbf{F} = (\gamma\mathcal{I}_{K}+2\mathbf{L}_{n})^{-1} $
We update the dual variables $\{\mathbf{\Lambda}_{n}\}$ via gradient ascent: 
\begin{equation}
[\mathbf{\Lambda}^{t}_{n}]^{k+1} = [\mathbf{\Lambda}^{t}_{n}]^{k} + \eta_{k}([\mathbf{D}^{t}_{n}]^{k}-\mathbf{B}\mathbf{diag}(\mathbf{c}_{n}))
\label{eqn:dual}
\end{equation}
We cycle through the primal-dual updates for $\{\mathbf{D}^{t}_{n}\}$ and $\{\mathbf{\Lambda}^{t}_{n}\}$ in Eq.~(\ref{eqn:primal}-\ref{eqn:dual}) to ensure that the constraints $\mathbf{D}^{t}_{n} = \mathbf{B} \mathbf{diag}(\mathbf{c}^{t}_{n})$ are satisfied with increasing certainty at each iteration. The learning rate parameter $\eta_{k}$ for the gradient ascent step is selected to a guarantee sufficient decrease in the objective for every iteration of alternating minimization. In practice, we initialize $\eta_{0}$ to $10^{-3}$, and scale it by $0.75$ at each iteration $k$.}

{\paragraph{\textbf{Step~{5}: Prediction on Unseen Data}}
In our cross-validated setting, we must compute the sr-DDL loadings $\{\mathbf{\Bar{c}}^{t}\}_{t=1}^{\bar{T}}$ for a new subject based on the $\mathbf{B}^{*}$ obtained from the training procedure and the new rs-fMRI correlation matrices $\{\bar{\mathbf{\Gamma}}^{t}\}$ and DTI Laplacians $\bar{\mathbf{L}}$. As we do not know the score $\mathbf{\bar{y}}$ for this individual, we need remove the contribution $\mathcal{L}(\cdot)$ from Eq.~(\ref{eqn:const}) and assume that the constraints $\bar{\mathbf{D}}^{t} = \mathbf{B^{*}}\mathbf{diag}(\bar{\mathbf{c}}^{t})$ are satisfied with equality. This effectively eliminates the Lagrangian terms. Essentially, the optimization for $\{\mathbf{\Bar{c}}^{t}\}$ now reduces to $\bar{T}_{n}$ decoupled quadratic programming (QP) objectives $\mathcal{Q}_{t}$: 
\begin{eqnarray*}
\Bar{\mathbf{c}}^{*t} =\argmin_{\bar{\mathbf{c}}^{t}}{\frac{1}{2}}~{(\mathbf{\Bar{c}}^{t})^{T}\mathbf{\Bar{H}}\mathbf{\Bar{c}}^{t}} + \mathbf{\Bar{f}}^{T}\mathbf{\Bar{c}}^{t} \ \ s.t.\ \  \mathbf{\Bar{A}}\mathbf{\Bar{c}}^{t} \leq \mathbf{\Bar{b}} \notag
\\ 
\mathbf{\Bar{H}} = 2(\mathbf{B^{*}}^{T}\bar{\mathbf{L}}\mathbf{B^{*}}); \ \ \ \ \ \ \ \ \ \ \ \ \ \ \nonumber \\ \ \ \ \ \ \ \ \ \mathbf{\Bar{f}} = -[\mathcal{I}_{K} \circ (\mathbf{B^{*}}^{T}(\bar{\mathbf{\Gamma}^{t}}\bar{\mathbf{L}}+ \bar{\mathbf{L}}\bar{\mathbf{\Gamma}^{t}})\mathbf{B^{*}})]\mathbf{1};   \ \ \ \ \ \ \ \ \ \ \ \ \ \   \\ \mathbf{\Bar{A}} = -\mathcal{I}_{K}  
 \   \mathbf{\Bar{b}} = \mathbf{0}  \ \ \ \ \ \ \ \ \ \ \ \ \ \ \
\end{eqnarray*}
Where $\circ$ is the elementwise Hadamard product. Notice that decoupling the objective across time allows us to parallelize this computation. Additionally, since $\bar{\mathbf{H}}$ is positive semi-definite, the formulation above is convex, leading to an efficient QP solution. Finally, we  estimate $\bar{\mathbf{y}}$ via a forward pass through the LSTM-ANN. }

\bibliography{mybibfile.bib}

\end{document}


\title{Supplemental Material for Deep sr-DDL: Deep Structurally Regularized Dynamic Dictionary Learning to Integrate Multimodal and Dynamic Functional Connectomics  data for Multidimensional Clinical Characterizations}

\author{\small{Niharika Shimona D'Souza}, \small{Mary Beth Nebel}, \small{Deana Crocetti}, \small{Nicholas Wymbs}, \\
\small{Joshua Robinson} , \small{Stewart Mostofsky}, \small{Archana Venkataraman}}
\date{}
\maketitle

\section{ Validation on Synthetic Data}
\label{AppdC}

\begin{figure}[b!]
   \centering
   \includegraphics[scale=0.6]{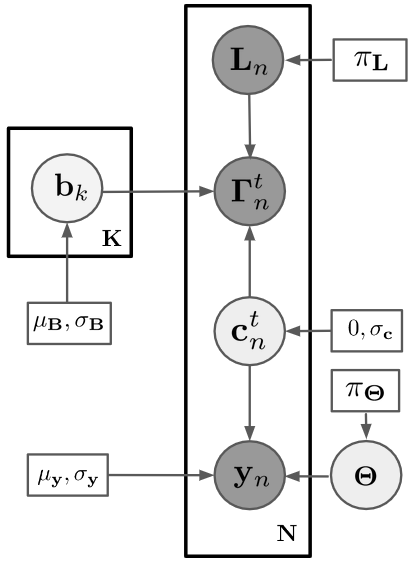}
   \caption{The graphical model for generating synthetic data. We fix the model parameters $\mathbf{\sigma}_{\mathbf{c}}= 4$, number of subjects $N$ at $60$, and number networks $K$ at $4$. The dimensionality of $\mathbf{y}_{n}$ is $M= 3$ and the length of the scan $T_n = 30$ for each subject. The shaded circles denote observed variables, while the clear circles indicate latent variables.}
   \label{fig:GraphicalModel}
\end{figure}

This experiment allows us to assess the behavior of our algorithm under various noise scenarios. The equivalent generating process for our framework is captured by the graphical model in Fig.~\ref{fig:GraphicalModel}.  As described in Section~2.2, the observed variables are the temporal correlation matrices $\{\mathbf{\Gamma}^{t}_{n}\}$, the DTI Laplacians $\mathbf{L}_{n}$, and the clinical scores $\{\mathbf{y}_{n}\}$, while the latent variables are the basis $\mathbf{B}$, the coefficients $\{\mathbf{c}^{t}_{n}\}$, and the neural network weights $\mathbf{\Theta}$.  Note that the dynamic correlation matrices $\{\mathbf{\Gamma}^{t}_{n}\}$ are completely described by the basis $\mathbf{B}$, the coefficients $\{\mathbf{c}^{t}_{n}\}$ and the Laplacian weighting $\mathbf{L}_{n}$. We further observe that the rs-fMRI data decompositions for each subject couple only through the shared basis and the clinical predictions through the shared network weights $\mathbf{\Theta}$. Conditioned on these variables, $\{\{\mathbf{\Gamma}^{t}_{n}\},\mathbf{L}_{n},\{\mathbf{c}^{t}_{n}\},\mathbf{\Theta},\mathbf{y}_{n}\}$  are independent across subjects. Fig.~\ref{fig:GraphicalModel} captures these conditional relationships.
\par We start by generating a basis matrix $\hat{\mathbf{B}} \in \mathcal{R}^{P \times K}$ by drawing its entries independently from a zero mean Gaussian with variance one. We then use the Gram-Schmidt procedure to compute an orthogonal basis $\mathbf{B}_{o} = \mathbf{orth}(\hat{\mathbf{B}})$. Finally, we simulate corruptions to this basis via additive Gaussian noise $\mathbf{B} = \mathbf{B}_{o} + \mathcal{N}(0,\sigma_{\mathbf{B}})$. Effectively, the value of $\sigma_{\mathbf{B}}$ quantifies the deviations of $\mathbf{B}$ from orthogonality, which is an assumption of our model. Note that the coefficient values in $\mathbf{c}_{n}$ are independent across networks and subjects, but not across time. Thus, for each subject, we generate the temporal coefficients using a isotropic Gaussian process with zero mean, and variance $\sigma_{\mathbf{c}}$. These values are clipped at $0$ to reflect the non-negativity in the coefficients. The variance parameter $\sigma_{\mathbf{c}}$ defines the scale of the coefficients. Next, we simulate the Graph Laplacians $\mathbf{L}_{n}$ for each subject based on structural connectivity priors computed using real-world data. Specifically, for each region pair, we first create a histogram of connectivity using binary adjacency matrices from the HCP database. With $\pi_{\mathbf{L}}$ denoting the probability of a connection between ROI pairs, we sample a symmetric graph adjacency matrix $\mathbf{A}_{n}$ per subject via a Bernouilli distribution with parameter $\pi_{\mathbf{L}}$. We then compute the corresponding Laplacians $\mathbf{L}_{n}$ from $\mathbf{A}_{n}$. This choice of prior helps us generate realistic structural connectivity profiles.  

\par Now, recall that our model seeks to approximate the rs-fMRI dynamic correlation matrices by $\mathbf{\Gamma}^{t}_{n} \approx \mathbf{B}\mathbf{diag}(\mathbf{c}^{t}_{n})\mathbf{B}^{T}$. Additionally, this decomposition is regularized by the individual Laplacians $\mathbf{L}_{n}$. Since we wish to evaluate the quality of this approximation, our generative model simulates $\mathbf{\Gamma}^{t}_{n}$ by adding structured noise (parametrized by $\mathbf{L}_{n}$) to $\mathbf{B}\mathbf{diag}(\mathbf{c}^{t}_{n})\mathbf{B}^{T}$. Specifically, we use the eigenbasis $\mathbf{X}$ of $\mathbf{L}_{n}$ to generate additive noise $\mathbf{N} = \sigma_{\mathbf{\Gamma}}\mathbf{X}\mathbf{X}^{T}$. We then compute the correlation matrices as $\mathbf{\Gamma}^{t}_{n} = \mathbf{B}\mathbf{diag}(\mathbf{c}^{t}_{n})\mathbf{B}^{T} + \mathbf{N}$. Note that this procedure preserves the positive semi-definiteness of the decomposition. Effectively, the parameter $\sigma_{\Gamma}$ controls the level of corruption in the observed dynamic correlation matrices. Finally, the observed variable $\{\mathbf{y}_{n}\}$, translates to a Gaussian with mean $\mathbf{\mu}_{\mathbf{y}_{n}}=\mathcal{F}_{\mathbf{\Theta}}(\{\mathbf{c}^{t}_{n}\}) \in \mathcal{R}^{M\times 1}$, and variance $\sigma_{\mathbf{y}_{n}}\mathbf{I}_{M}$. The function mapping $\mathcal{F}_{\mathbf{\Theta}}$ refers to the LSTM-ANN network with the parameters $\mathbf{\Theta}$ - which we randomly initialize. This is again folded to reflect positive values of $\mathbf{y}_{n}$. Here, $\sigma_{\mathbf{y}}$ controls the noise in the clinical scores.

\begin{figure*}[t!]
   \centering
   \fbox{\includegraphics[width= \textwidth-5mm]{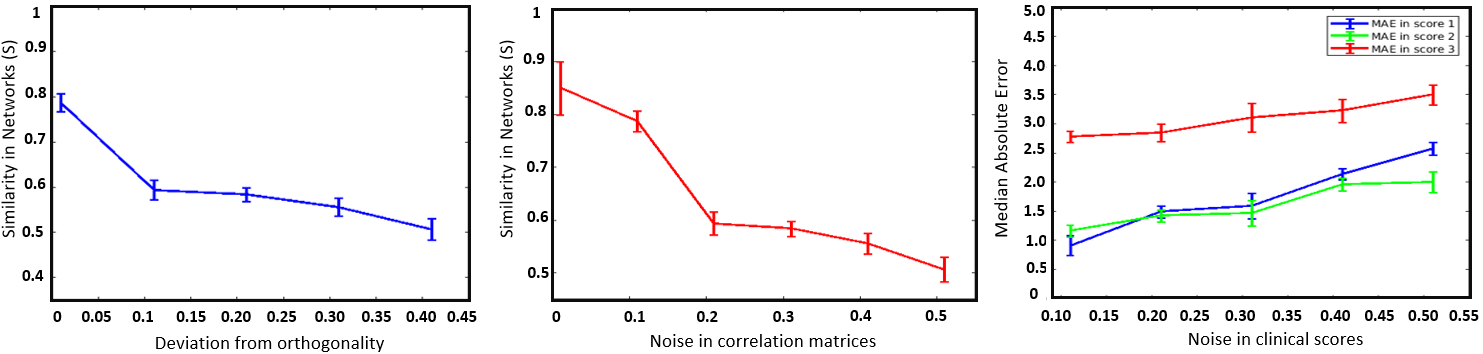}}
 \footnotesize{  \caption{Performance on synthetic experiments. (\textbf{L}): Varying the level of deviation from orthogonality  ($\mathbf{\sigma}_{\mathbf{\Gamma}}=0.2$, $\mathbf{\sigma}_{\mathbf{Y}} = 0.2$), (\textbf{M}): Varying the level of noise in $\mathbf{\Gamma}$ ($\mathbf{\sigma}_{\mathbf{B}} =0.2$, $\mathbf{\sigma}_{\mathbf{y}}=0.2$) , (\textbf{R}): Varying the level of noise in $\mathbf{y}_{n}$ under ($\mathbf{\sigma}_{\mathbf{B}} =0.2$, $\mathbf{\sigma}_{\mathbf{\Gamma}} = 0.2$) Values on the x-axis have been normalized to reflect a $[0-1]$ range by dividing by the maximum value of the variable. We report deviations from the mean for recovered similarity/MAE at each parameter setting in terms of a standard error value. The reported $x$-axis range reflects the regimes within which the algorithm converges to a local solution }}
   \label{fig:Simulated}
\end{figure*}

\par There are two sources of noise for the observed variables. The first is error in the correlation matrices $\mathbf{\Gamma}^{t}_{n}$, controlled by changing $\sigma_{\mathbf{\Gamma}}$. The second case is error in the clinical scores $\mathbf{y}_{n}$, quantified by the parameter $\sigma_{\mathbf{y}}$. Additionally, we are also interested in evaluating the performance under varying levels of deviations of the basis from orthogonality. This is controlled by the parameter $\sigma_{\mathbf{B}}$.

\par We evaluate the efficacy of our algorithm using two separate metrics. {The first is an average absolute cosine similarity measure $S$ between each recovered network, $\mathbf{\bar{b}}_{k}$, and its corresponding best matched ground truth network,} $\mathbf{b}_{k}$, normalizing the latter to unit norm, that is:
\begin{equation}
{S} ={{\frac{1}{K}}\sum_{k}\frac{\vert{ \mathbf{b}_{k}^{T}  \mathbf{\bar{b}}_{k}}\vert}{{\vert\vert{\mathbf{b}_{k}}\vert\vert}_{2}}}{.}
\label{eqn:similarity}
\end{equation}
The second metric is the Median Absolute Error (MAE) between the output of the trained LSTM-ANN $\hat{\mathbf{y}}_{n}$ and the true scores $\mathbf{y}_{n}$.

\par Fig.~\ref{fig:Simulated} depicts the performance of the algorithm in these three cases. In the each subplots, the $x$-axis corresponds to increasing the levels of noise. In the first two subplots, the $y$-axis indicates the similarity metric $S$ computed for the particular setting, while in the rightmost subplot, we plot the MAE for predicting the three scores. All numerical results have been aggregated over $50$ independent trials.

\par In the leftmost plot, an $x$-axis value close to $0$ indicates low levels of deviation of $\mathbf{B}$ from orthogonality, while increasing values corresponds to a more severe deviation from the modeling assumptions. During this experiment, the values of the other free parameters in Fig.~\ref{fig:GraphicalModel} were held constant. We observed that the MAE of the three scores remains roughly constant for all noise settings (score $1$\textemdash$1.49~\rpm~0.09$, score $2$\textemdash$1.34~\rpm~0.07$, score $3$\textemdash$3.10~\rpm~0.11$). The middle plot evaluates subnetwork recovery when the noise in the dynamic correlation matrices, i.e. $\mathbf{\sigma}_{\mathbf{\Gamma}}$ is increased. The $\mathbf{x}$-axis reports normalized values of $\mathbf{\sigma}_{\mathbf{\Gamma}_{n}}$ while the remaining free parameters were held constant. Similar to the previous scenario, the MAE remains roughly constant for varying noise settings (score $1$\textemdash$1.50~\rpm~0.08$, score $2$\textemdash$1.50~\rpm~0.06$, score $3$\textemdash$2.96~\rpm~0.50$). Finally, the rightmost plot in Fig.~\ref{fig:Simulated} indicates performance under varying noise in the scores $\mathbf{y}_{n}$. Again, normalized $\mathbf{\sigma}_{\mathbf{y}}$ values are reported on the x-axis. For this experiment, we observed that $S = 0.87~\rpm~0.05$ for varying noise levels. 

\par As expected, increased noise in the correlation matrices and deviations from orthogonality worsens recovery performance of the algorithm. This is reflected by the decay in the similarity measure along with increasing noise parameters. Since the parameter $\sigma_{\mathbf{y}}$ is held constant, we do not observe much variation in the the MAE values upon increasing the noise. Lastly, we notice that the algorithm performs better when the level of noise in the scores is lower. This is indicated by the increasing values of MAE in the right subplot in Fig.~\ref{fig:Simulated}. Since $\sigma_{\mathbf{B}}$ is held constant for this experiment, the metric $S$ remains fairly constant even upon increasing the noise in the scores.

\par Taken together, our simulations indicate that the optimization procedure is robust in the noise regime ($0.01-0.2$) estimated from the real-world rs-fMRI data. In addition, these experiments help us identify the stable parameter settings ($\lambda = 1-10$, learning rates) which govern the convergence of the algorithm which guide our real world experiments.